\documentclass[pdflatex,sn-nature]{sn-jnl}

\usepackage{graphicx}%
\usepackage{multirow}%
\usepackage{amsmath,amssymb,amsfonts}%
\usepackage{amsthm}%
\usepackage{mathrsfs}%
\usepackage[title]{appendix}%
\usepackage{xcolor}%
\usepackage{textcomp}%
\usepackage{manyfoot}%
\usepackage{booktabs}%
\usepackage{algorithm}%
\usepackage{algorithmicx}%
\usepackage{algpseudocode}%
\usepackage{listings}%

\usepackage{graphicx}
\usepackage{amsmath}
\usepackage{amsfonts}
\usepackage{bm}
\usepackage{xr}
\usepackage{blindtext}
\usepackage{times}
\usepackage{subcaption}
\usepackage{prettyref}
\usepackage{algpseudocode}
\usepackage{lineno}

\usepackage{xcolor}
\usepackage{tcolorbox}

\usepackage[inline,shortlabels]{enumitem}

\usepackage{hyperref}
\hypersetup{
    colorlinks,
    citecolor=black,
    filecolor=black,
    linkcolor=black,
    urlcolor=blue
}

\definecolor{red}{HTML}{ef606f}
\definecolor{blue}{HTML}{b1c1e6}
\definecolor{yellow}{HTML}{ffd15f}

\newcommand{\figrange}[2]{Fig.~\ref{#1}--\ref{#2}}
\renewcommand{\figurename}{Fig.}
\newrefformat{fig}{\hyperref[#1]{Fig.~\ref*{#1}}}
\newrefformat{suppfig}{\hyperref[#1]{Supplementary Fig.~\ref*{#1}}}
\newrefformat{eq}{Eq.~\ref{#1}}
\newrefformat{alg}{Algorithm~\ref{#1}}
\newrefformat{sec}{\nameref{#1}}

\algnewcommand{\Inputs}[1]{%
  \State \textbf{Inputs:}
  \Statex \hspace*{\algorithmicindent}\parbox[t]{.8\linewidth}{\raggedright #1}
}
\algnewcommand{\Outputs}[1]{%
  \State \textbf{Outputs:}
  \Statex \hspace*{\algorithmicindent}\parbox[t]{.8\linewidth}{\raggedright #1}
}
\algnewcommand{\Initialize}[1]{%
  \State \textbf{Initialize:}
  \Statex \hspace*{\algorithmicindent}\parbox[t]{.8\linewidth}{\raggedright #1}
}

\theoremstyle{thmstyleone}%
\theoremstyle{thmstyletwo}%

\theoremstyle{thmstylethree}%

\begin{document}

\title[Article Title]{A computational framework for understanding embodied plans}

\author[1]{\fnm{Aalap D.} \sur{Shah}}\email{aalap.shah@yale.edu}
\author[4]{\fnm{Basil} \sur{Saeed}}\email{bsaeed@stanford.edu}
\author[5]{\fnm{Joshua B.} \sur{Tenenbaum}}\email{jbt@mit.edu}
\author[3]{\fnm{Tobias} \sur{Gerstenberg}}\email{gerstenberg@stanford.edu}
\author[3]{\fnm{Hyowon} \sur{Gweon}}\email{hyo@stanford.edu}
\author*[1,2]{\fnm{Ilker} \sur{Yildirim}}\email{ilker.yildirim@yale.edu}

\affil*[1]{\orgdiv{Department of Psychology}, \orgname{Yale University}, \orgaddress{\city{New Haven}, \state{CT}, \country{USA}}}
\affil[2]{\orgdiv{Wu Tsai Institute}, \orgname{Yale University}, \orgaddress{\city{New Haven}, \state{CT}, \country{USA}}}
\affil[3]{\orgdiv{Department of Psychology}, \orgname{Stanford University}, \orgaddress{\city{Palo Alto}, \state{CA}, \country{USA}}}
\affil[4]{\orgdiv{Department of Statistics}, \orgname{Stanford University}, \orgaddress{\city{Palo Alto}, \state{CA}, \country{USA}}}
\affil[5]{\orgdiv{Department of Brain \& Cognitive Sciences}, \orgname{MIT}, \orgaddress{\city{Cambridge}, \state{MA}, \country{USA}}}


\abstract{Reasoning about people's plans and goals based on their actions is fundamental to human social interaction and essential for cooperative, embodied artificial intelligence. Formalizing this capacity in naturalistic, embodied settings has been a longstanding challenge. Here, we present GenEA, a novel agent representation capable of reasoning over multi-step, long-horizon embodied plans. At the heart of GenEA lies a multilevel state-space that couples the coarse biomechanics of an articulated body and its interactions with nearby objects by extending the task-and-motion planning framework from robotics with simulation-based intuitive physics. By embedding this state-space in a generative model, GenEA yields a joint probability distribution over goals and embodied plans. Across several experiments spanning the qualitatively distinct domains of object manipulation and navigation, featuring diverse tasks and multiple behavioral measures, we show that GenEA, through simple transformations of this joint distribution, recapitulates how humans reason about embodied plans. Crucially, GenEA consistently outperforms strong alternative models, including plausible heuristics, standard task-and-motion planning, and a state-of-the-art vision-language model. GenEA thereby offers a new perspective on the architecture of human reasoning and social cognition, suggesting that embodied artificial intelligence systems seeking to seamlessly interact with humans should adopt GenEA-like generative models of action planning.}


\maketitle

\clearpage
\section*{Introduction}

\begin{figure*}[t]
\centering
\includegraphics[width=\linewidth]{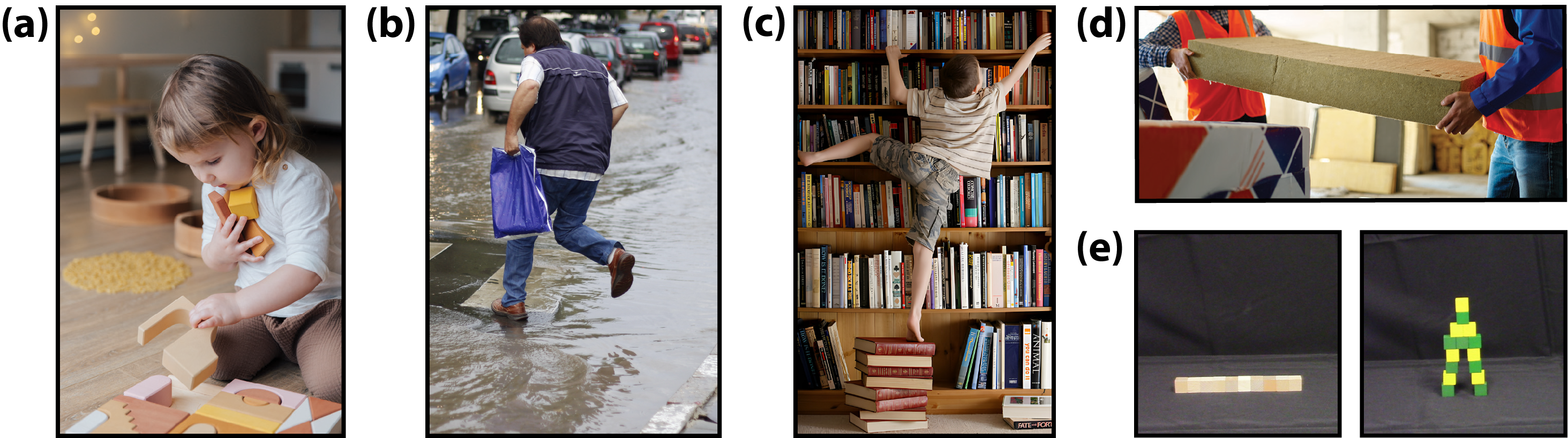}
\caption{ \textbf{Understanding embodied plans.}
\textbf{(a)} We can tell that the child's goal location for slotting the block held in her left hand is the space beside the pink block, and can reason about why she might be holding onto several other pieces between her chin and right hand. 
\textbf{(b)} We can appreciate that the man's jump helps him simultaneously achieve several goals, one being clearing the puddle.
Reasoning about others' embodied plans is critical for our everyday physical and social interactions, including knowing when to intervene to safeguard a child or a friend \textbf{(c)}, collaborating with others to form and accomplish shared embodied goals \textbf{(d)}, 
and making informed decisions about which tasks to undertake by considering task difficulty relative to our own skill level \textbf{(e)}.
Images (a-d) were taken either from stock.adobe.com or Freepik. Images in (e) were borrowed from Gweon et al. \cite{gweon2017reverse}.}
\begingroup
\phantomsubcaption
\label{fig:introA}
\phantomsubcaption
\label{fig:introB}
\phantomsubcaption
\label{fig:introC}
\phantomsubcaption
\label{fig:introD}
\phantomsubcaption
\label{fig:introE}
\endgroup
\label{fig:intro}
\end{figure*}

A close inspection of \prettyref{fig:introA}, showcasing a seemingly simple scenario of a child playing with blocks, reveals a rich understanding of the child's goals through the nuances of her actions. We can tell that her left-hand action is a goal-driven maneuver intended to place the block from her hand next to the pink block. Her posture, bracing several pieces against her chin, appears effortful, inviting further inferences about her plans: perhaps she already tested these blocks and found they did not fit, or maybe she intends to use them next. Likewise, in \prettyref{fig:introB}, we see a man executing a jump that appears to accomplish several goals at once: clearing the puddle without splashing, safeguarding the item in his hand, and securing a stable landing. 

For decades, philosophers and cognitive scientists have sought to characterize this ability to reason about the rational, purposeful actions of others (e.g., \citenum{dennett1989intentional,gergely2003teleological,wilson2001perceiving,sebanz2006joint,goldman2009social,baker2009action,jara2016naive,jara2020naive}), an ability critical for everyday social and physical interaction, such as intervening to safeguard a child (\prettyref{fig:introC}), coordinating with others (\prettyref{fig:introD}) or determining which tasks are appropriately matched, in terms of difficulty, to one's skill level (\prettyref{fig:introE}). Underlying much of this work is the pursuit of identifying an ``agent representation'' that supports rational inference over an agent's actions and goals, given its local environment. Existing accounts \cite{baker2009action,jara2016naive,baker2017rational,jara2020naive,ullman2009help,wu2023computational,liu2017ten,shu2021agent,berke2023thinking,ying2026grounding} have abstracted agents into point-like entities operating within two-dimensional grid-world-like environments governed by highly constrained transition rules for agents' actions and the effects of those actions on the environment. The open-ended and dynamic nature of naturalistic, embodied settings (as in \figrange{fig:introA}{fig:introE}), however, presents a qualitatively distinct and quantitatively greater challenge \cite{nuzzi2026continuous}. Take, for example, the task of helping someone navigate to a specific landmark. The route can be described as a sequence of symbolic actions (e.g., turn left, move forward) paired with tokens of other relevant landmarks en route. For this task, representing the person as a point-like entity alone would suffice: the resulting plan would be no different than if their body had been modeled explicitly. Instead, consider the same task, but for a person in a wheelchair. To account for variations in the terrain and other accessibility constraints, path planning now demands that the agent's body and its local environment be treated as integral to the plan itself. This highlights that the agent representation in the mind is fundamentally richer, and perhaps structured differently, than any low-dimensional, symbolic state space can accommodate.

So here, we introduce the Generative Embodied Agent (GenEA), a novel, multilevel agent representation whose state space couples the coarse biomechanics of an articulated body and its interactions with nearby objects to enable reasoning over multi-step, long-horizon action sequences. GenEA synthesizes core principles from biological motor control (feedback control and noise tolerance) \cite{mcnamee2019internal} to extend the two-level abstraction of the task-and-motion planning framework \cite{cambon2009hybrid,kaelbling2011hierarchical,srivastava2014combined,garrett2021integrated} from robotics, composed of the symbolic and kinematics levels, with a dynamics level that implements intuitive physics \cite{battaglia2013simulation}. GenEA embeds this state space into a generative model of action planning, yielding a joint probability distribution over rewards and embodied plans that captures the constraints and trade-offs inherent in naturalistic settings. Reasoning about diverse tasks, such as drawing inferences about the child's effortful posture or the man's calculated jump, then emerges as simple transformations of this joint distribution. We evaluated GenEA's performance against that of humans and alternatives, including an ablation, heuristics baselines, and a popular frontier vision-language model as an example of the large-scale statistical learning approach. Crucially, we did so across two fundamentally distinct domains of manipulation, involving reasoning about block structures, and navigation, involving reasoning about obstacle-course settings. We found that, across both domains, GenEA's performance closely aligned with human judgments and, where applicable, response times, while substantially outperforming alternative models. In doing so, GenEA motivates a new perspective on the architecture of human reasoning and social cognition.

\begin{figure}[b!]
\centering
\includegraphics[width=\linewidth]{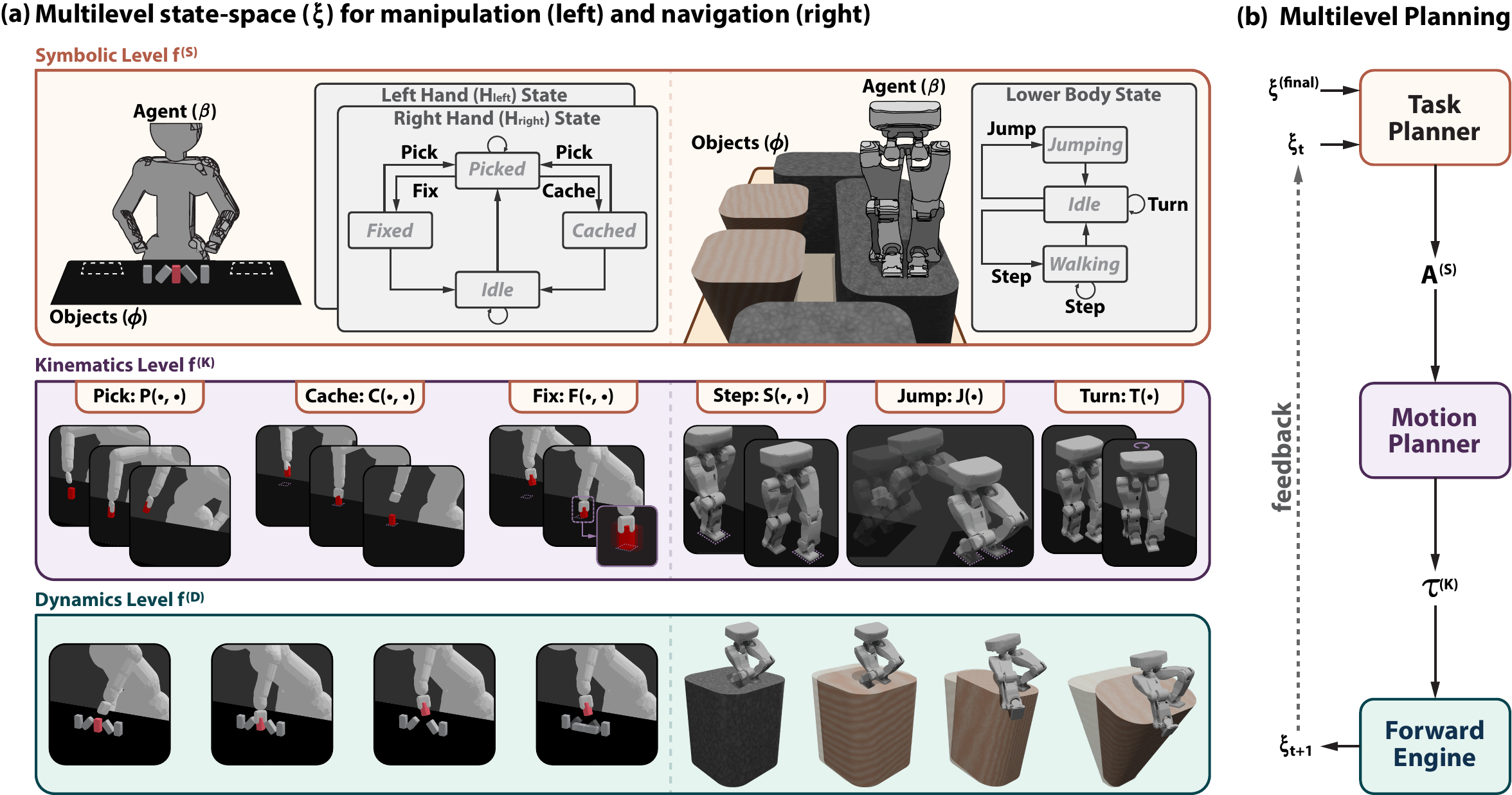}
\caption{\textbf{Overview of GenEA.} \textbf{(a)} At the core of GenEA is a multilevel state-space that couples the coarse biomechanics of an agent $\beta$ and its interactions with the nearby objects $\phi$. 
Its Symbolic Level $f^{(\text{S})}$ defines an automaton of action primitives for the body (such as picking up an object or jumping across platforms), and how these actions impact the objects in the scene. 
The Kinematics Level $f^{(\text{K})}$ captures the noisy motion trajectories of the agent's end-effectors and, correspondingly, of objects being manipulated. 
The Dynamics Level $f^{(\text{D})}$ updates the objects' states by simulating gravity and the forces acting between them.
\textbf{(b)} GenEA defines a conditional distribution over plans under this multilevel state space. Given a reward, a specified final configuration, GenEA performs classical planning using a task planner, a motion planner, and a forward engine (each corresponding to the three functions of the multilevel state space). 
}
\begingroup
\phantomsubcaption
\label{fig:computational_modelA}
\phantomsubcaption
\label{fig:computational_modelB}
\phantomsubcaption
\label{fig:computational_modelC}
\endgroup
\label{fig:computational_model}
\end{figure}

\section*{The generative embodied agent (GenEA) model}

GenEA provides a joint probability distribution $Pr(\pi,\,R\,|\,\xi)$ over embodied plans $\pi$ and rewards $R$, factorized as
\begin{align}
Pr(\pi,\,R\,|\,\xi) = Pr(R\,|\,\xi)\,Pr(\pi\,|\,R,\,\xi)
\label{eq:eq1}
\end{align}
\noindent where $\xi$ represents the coupled multilevel state-space (see below) of an agent and the nearby physical objects, $Pr(R\,|\,\xi)$ is a distribution over possible rewards (defined as the final configuration that the agent must achieve), and $Pr(\pi\,|\,R,\,\xi)$ is a distribution over embodied plans. As we show below, this formulation has remarkable, and, relative to the alternatives we tested, unique explanatory power for how humans reason about these embodied plans.

\subsection*{A multilevel state-space, and planning within it}

GenEA makes \prettyref{eq:eq1} computable via a multilevel state-space 
$\xi = \{\beta,\, \phi,\, f^{(\text{S})},\, f^{(\text{K})},\, f^{(\text{D})}\}$ (\prettyref{fig:computational_modelA}) which specifies $\beta$, the main degrees-of-freedom of the task-relevant part of the human body, and $phi$, an array of indexable 3D objects along with their positions, orientations and physical properties (such as, mass). The changes in the states of the agent and objects are structured into three levels --- the symbolic level, $f^{(\text{S})}$; the kinematics level, $f^{(\text{K})}$; and the dynamics level, $f^{(\text{D})}$. Planning within this multilevel state-space defines the conditional distribution over embodied plans Pr$(\pi\,|\,R,\,\xi)$ in \prettyref{eq:eq1}. As an account of how people reason about embodied plans, we posit that this multilevel agent representation captures key aspects of how the mind reasons about embodied plans, enabling quantitative comparisons between GenEA and human judgments, as well as response times where applicable, across a range of reasoning queries.

\subsubsection*{Symbolic level, $f^{(\text{S})}$}

This level defines a set of symbolic actions $\text{a}^{(\text{S})} \in \text{A}^{(\text{S})}$ sufficient to perform a given task, while abstracting away the details of how these actions are executed or how they might affect object trajectories. The symbolic actions available are therefore based on the body parts primarily involved in the task. For tasks involving the upper body, such as block structure building, the symbolic action set includes picking up a block \textproc{Pick(Hand, $\phi$)}, temporarily placing a block on the table \textproc{Cache(Hand, Location)}, and fixing a block at its final location \textproc{Fix(Hand, Location)}. For tasks involving the lower body, such as navigation, the symbolic action set includes walking on or between platforms \textproc{Step(Length, Count)}, jumping between platforms \textproc{Jump(Location)}, and turning in place \textproc{Turn(Angle)}. Although we predefine these action primitives based on widely adopted use cases in robotics, the action dictionary can be extended using recent learning-based approaches for symbolic-level planning (e.g., \cite{garrett2021integrated, li2025bilevel}). We also define a symbolic automaton (\prettyref{fig:computational_modelA}, uppermost panel) that governs the transition rules between these actions --- $f^{(\text{S})}: \{\beta,\, \phi\} \rightarrow \text{A}^{(\text{S})}$. 

\subsubsection*{Kinematics level, $f^{(\text{K})}$}

The kinematics level translates abstract symbolic actions into executable action trajectories. Each symbolic action specifies a start and end point from which a trajectory is defined: a series of 3D points delineating the desired path between the endpoints, $\bm{\tau}^{(\text{K})} = \{\tau^{(\text{K})}\}_j$ for each joint $j$ in the body (\prettyref{fig:computational_modelA}, middle panel). Note that if the symbolic action involves manipulating an object, for instance picking it up, the object's trajectory is constrained by the trajectory of the end effector responsible for the action. We denote the kinematics level as $f^{(\text{K})}: \{\beta,\, \phi,\, \text{a}^{(\text{S})}\} \rightarrow \bm{\tau^{(\text{K})}}$.

Thus far, the formulations of the symbolic and kinematics levels directly follow the standard task-and-motion planning (TAMP) framework (specifically, \citenum{toussaint2015logic}). We denote this two-level state-space of TAMP as $\xi^{(\text{TAMP})}=\{\beta,\, \phi,\, f^{(\text{S})},\, f^{(\text{K})}\}$. TAMP remains the state-of-the-art approach in long-horizon robotics frameworks \cite{garrett2021integrated} and TAMP-like planning, which focuses on posture-based constraints and kinematics, has been used successfully to model basic human manual control \cite{rosenbaum2001posture}.

GenEA extends the standard TAMP framework to include kinematic uncertainty, treating each trajectory endpoint as noisy rather than fixed. Including kinematic uncertainty avoids building in any assumptions of unrealistic precision, thereby supporting more robust planning, and, importantly, provides a principled interface through which the dynamics level (described below) can be incorporated into the state space. Kinematic uncertainty was implemented specifically as part of the \textproc{Fix}, \textproc{Step} and \textproc{Jump} actions, whose endpoints were modeled using a truncated skewed normal distribution. The kinematic updates to the state of the body and objects in GenEA is given by $\{\beta,\,\phi\} \xrightarrow{\bm{\tau}^{(\text{K})} + \epsilon} \{\beta^{(\text{K})},\,\phi^{(\text{K})}\}$.

\subsubsection*{Dynamics level, $f^{(\text{D})}$}
The dynamics level, unique to GenEA, incorporates knowledge of how objects move and interact with each other under the influence of gravity. Inspired by earlier modeling work in physical reasoning \cite{battaglia2013simulation}, GenEA updates the objects' poses (position and orientation) (\prettyref{fig:computational_modelA}, lowermost panel) using game-engine style, approximate but efficient simulations --- $f^{(\text{D})}: \phi^{(\text{K})} \rightarrow \phi^{(\text{D})}$ (for details, see \prettyref{sec:methods}). As a consequence of applying this physical simulation, the actions taken by the agent may result in advancing or hindering its pursuit of the reward (final) configuration. For instance, the action of picking up the red block, depicted in \prettyref{fig:computational_model_adjA}, causes the blocks it previously supported to fall into their respective final positions due to gravity.

\subsubsection*{A distribution over embodied plans}

We define the conditional distribution over plans, Pr$(\pi\,|\,R,\, \xi)$, by planning within the multilevel state-space $\xi$. The planner accepts as inputs a reward, specified by the final object configuration $\phi^{(\text{final})}$, and the initial object configuration $\phi^{(\text{init})}$. It then searches over $\xi$, transitioning through the symbolic, kinematics, and dynamics levels in a feedback-like loop (\prettyref{fig:computational_modelB}) to generate embodied plans that achieve the specified reward. 

We denote a plan as $\pi=\{\xi_1,\,a_1,\,...,\,\xi_{T-1},\,a_{T-1},\,\xi_T\}$, where $\phi^{(\text{init})} \in \xi_1$, $\phi^{(\text{final})}\in \xi_T$, and each action $a_t$ encompasses a symbolic action, the kinematic trajectories that accomplish this symbolic action, and the dynamics-updated object states. The probability of these plans is given by the Boltzmann distribution $Pr(\pi\,|\,R,\, \xi) \propto \exp(-\kappa \, C_{\xi}(\pi))$, where $C$ represents the cost of executing the plan and $\kappa$ is the temperature parameter. (Note that our results are not sensitive to the specific choice of this parameter, see \prettyref{suppfig:supp_temp_prob_pearson}.) This formulation establishes an inverse relationship between probability and cost, indicating that the probability of a plan decreases exponentially as its cost increases, thereby favoring more efficient plans. For this reason, the conditional $Pr(\pi\,|\,R,\, \xi)$  reflects the intuitive psychological principle that agents tend to take actions that maximize rewards and minimize costs \cite{jara2016naive}.

While various cost functions could in principle be adopted, here we report results based on a straightforward choice: the total kinetic energy incurred by executing the kinematic trajectories in the forward engine. Our results --- i.e., GenEA's ability to better explain human behavior relative to alternative models across all tasks and settings --- are not sensitive to this particular choice of cost functions. That is, even under other cost functions, including the ``plan length'' and ``solver-derived optimization costs'' (either ``sum of squared accelerations'' or ``sum of squared residuals''), GenEA continues to outperform alternative models (see \prettyref{suppfig:supp_cost_funcs}). Kinetic energy is nonetheless particularly appealing not only for its simplicity, but also for its direct relationship to embodiment research \cite{rosenbaum2001posture}. In \prettyref{suppfig:supp_temp_prob_pearson}, we provide evidence that no other cost function outperforms kinetic energy.

\begin{figure}[b!]
\centering
\includegraphics[width=.75\linewidth]{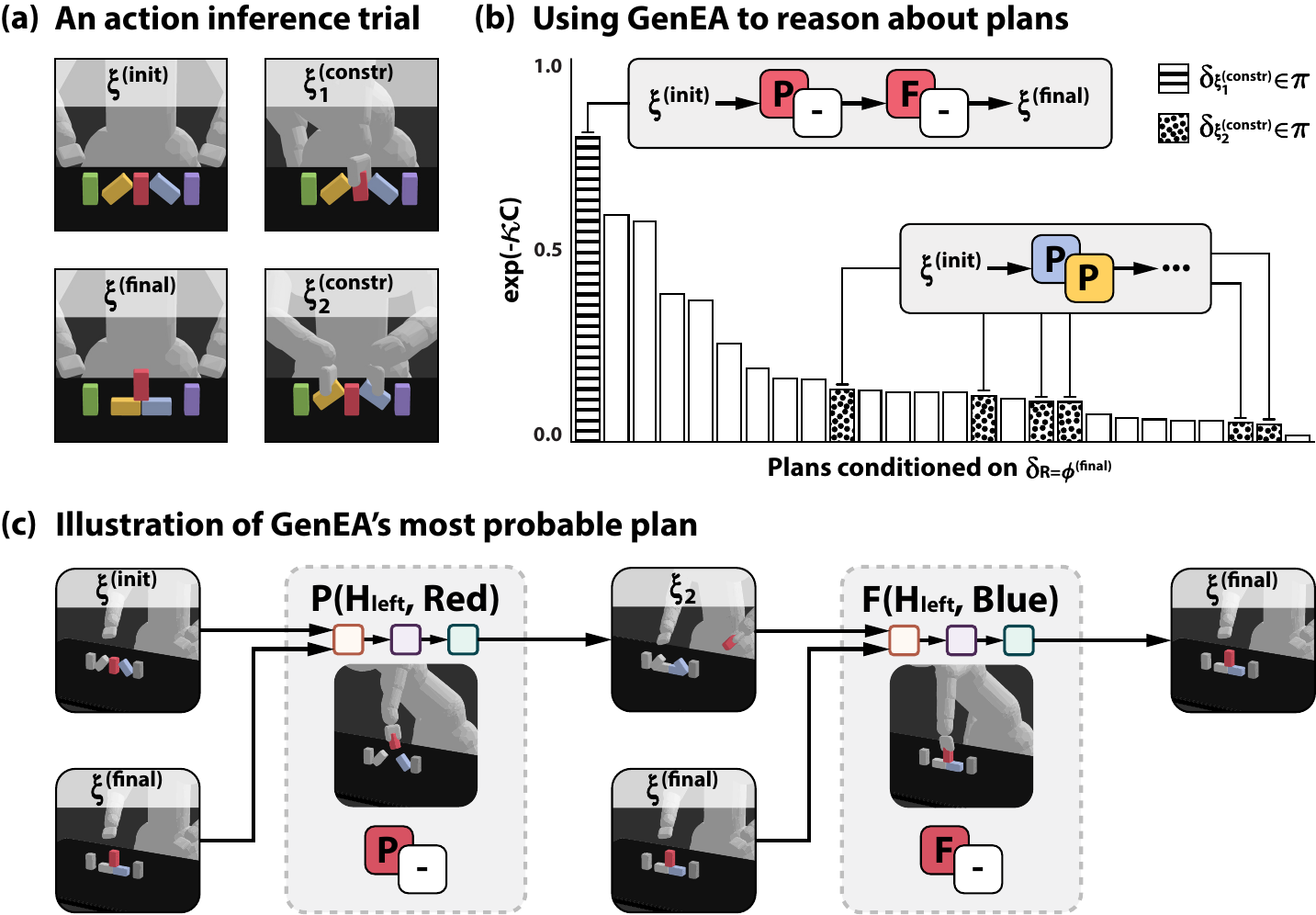}
\caption{\textbf{Reasoning queries in GenEA}
\textbf{(a)} An example action inference trial where the GenEA model is provided with the reward query $\phi^{(\text{final})} \in \xi^{(\text{final})}$ and two plan constraints: $\xi^{(\text{constr})}_1$ representing a one-handed \textproc{Pick} action on the red block \tcbox[colback=red, colframe=black, arc=1pt, boxrule=1pt, on line, boxsep=.5pt, left=1pt, right=1pt, top=1pt, bottom=1pt]{\textcolor{black}{\textsf{\textbf{P}}}} (right hand is idle), and $\xi^{(\text{constr})}_2$ representing a two-handed \textproc{Pick} action where the left hand picks the blue block \tcbox[colback=blue, colframe=black, arc=1pt, boxrule=1pt, on line, boxsep=.5pt, left=1pt, right=1pt, top=1pt, bottom=1pt]{\textcolor{black}{\textsf{\textbf{P}}}} and a right hand picks the yellow block \tcbox[colback=yellow, colframe=black, arc=1pt, boxrule=1pt, on line, boxsep=.5pt, left=1pt, right=1pt, top=1pt, bottom=1pt]{\textcolor{black}{\textsf{\textbf{P}}}}. The model is also given access to the initial configuration $\phi^{(\text{init})} \in \xi^{(\text{init})}$.
\textbf{(b)} A bar plot showing the distribution of plans based on the reward query. Each bar corresponds to a different plan, with textured bars highlighting plans that further satisfy the corresponding plan-constraint query.
\textbf{(c)} An illustration of a particularly efficient plan, generated by GenEA, that takes advantage of gravity in achieving the reward configuration. This plan consists of two one-handed actions. The first left-handed action, to \textproc{Pick} the red block, is represented as a red box \tcbox[colback=red, colframe=black, arc=1pt, boxrule=1pt, on line, boxsep=.5pt, left=1pt, right=1pt, top=1pt, bottom=1pt]{\textcolor{black}{\textsf{\textbf{P}}}} paired with a white box marked with a dash indicating that the right hand is idle. Similarly, the second action is depicted as a red box \tcbox[colback=red, colframe=black, arc=1pt, boxrule=1pt, on line, boxsep=.5pt, left=1pt, right=1pt, top=1pt, bottom=1pt]{\textcolor{black}{\textsf{\textbf{F}}}} (denoting a left-handed \textproc{Fix} action to position the red block relative to the blue block), alongside a white box with a dash (denoting the idle right hand)
}
\begingroup
\phantomsubcaption
\label{fig:computational_model_adjA}
\phantomsubcaption
\label{fig:computational_model_adjB}
\phantomsubcaption
\label{fig:computational_model_adjC}
\endgroup
\label{fig:computational_model_adj}
\end{figure}

\prettyref{fig:computational_model_adjA} shows an example action inference trial, \prettyref{fig:computational_model_adjB} its distribution over plans and \prettyref{fig:computational_model_adjC} the most probable plan. For further details of our planner, please refer to \prettyref{sec:methods}.

\subsection*{Using GenEA to reason about plans}

GenEA takes as input a query $Q$, which formalizes a given reasoning task as summaries of the joint distribution in \prettyref{eq:eq1}. A query consists of two components $Q=\{Q_R,\,Q_{\pi}\}$. $Q_R = \{\phi^{(\text{final})}_i\}$, represents a set of $i~\geq~1$ reward configurations, and $Q_\pi = \{\xi^{(\text{constr})}_j\}$ is a set of $j~\geq~1$ constraints over plans. In the absence of any plan constraints, $Q_\pi = \emptyset$. Consider, for instance, the action inference task in \prettyref{fig:computational_model_adjA}. In this task, GenEA would be provided with $Q_R=\{\phi^{(\text{final})}\}$ and the plan constraints $Q_\pi=\{\xi^{(\text{constr})}_1,\,\xi^{(\text{constr})}_2\}$, provided individually. 

In addition to the query $Q$, GenEA receives the initial object configuration, $\phi^{(\text{init})}$. Together, they encapsulate all the task-relevant information available to human participants in our experiments, thereby enabling GenEA to reason about plans for the same tasks that humans encounter.

To solve a query, we calculate the probability vector of achieving desired reward configurations ($Q_R$) while also satisfying plan constraints ($Q_{\pi}$):
\begin{align}
\vec{v}_{i,\,j} = Pr(\pi,\,R\,|\,\xi) \cdot \delta_{R = \phi^{(\text{final})}_i}\cdot\delta_{\xi^{(\text{constr})}_j \in \pi}
\label{eq:eq2}
\end{align}
\noindent where the Kronecker delta function $\delta_{R = \phi^{(\text{final})}_i}$ is 1 if $R$ is the desired final configuration (i.e., $\phi^{(\text{final})}$) and 0 otherwise, and $\delta_{\xi^{(\text{constr})}_i \in \pi}$ is 1 if the state $\xi^{(\text{constr})}_i$ is visited by $\pi$ and 0 otherwise. Note that $\emptyset \in \pi$ is true or ``1'' for all plans, meaning that when $Q_\pi=\emptyset$, the evaluation solely depends on $Q_R$. 
Because Pr$(\pi,\,R\,|\,\xi)\cdot \delta_{R = \phi^{(\text{final})}}$, which selects plans that arrive at $\phi^{(\text{final})}$, is equivalent to the conditional Pr$(\pi\,|\, R = \phi^{(\text{final})}, \xi)$, $\vec{v}_{i,\,j}$  can be computed efficiently by considering only the distribution of plans satisfying the queried reward configuration (\prettyref{fig:computational_model_adjB}).
To make a decision across a set of queries (e.g., selecting among alternative actions in \prettyref{fig:computational_model_adjA}), we use a simple, parameter-free choice rule $\max (\vec{v}_{i,\,j})/\sum_{i^\prime}\sum_{j^\prime}{\max( \vec{v}_{i^\prime,\,j^\prime}})$, where $(i^\prime,\,j^\prime)$ spans all possible options in a given reasoning task, representing the model's preference for the query set $(i,\,j)$. 

\section*{Alternative agent representations}

We tested GenEA against a range of alternative models, including task-specific heuristic models, the TAMP ablation, and a state-of-the-art vision-language model. 

The TAMP model, as described above, is also a generative model for action planning, parameterizing a joint distribution over rewards and plans using the same cost function as GenEA. However, crucially, it solely operates within the standard task-and-motion planning state-space, $\xi^{(\text{TAMP})}$, instead of GenEA's $\xi$. Comparing GenEA to the TAMP model therefore allows us to isolate the specific contribution of intuitive physics --- captured by the dynamics level, which TAMP excludes --- to how humans reason about embodied plans.

The task-specific heuristic models, which we designed separately for each experiment as part of our pre-registrations \cite{shah2022,shah2023,shah2026}, challenge some of the fundamental assumptions we made in GenEA. Specifically, these models explore the possibility that reasoning about embodied actions does not rely on a generative model of action planning since they have no representation of the agent's body or its local environment. Instead, the heuristic models provide approximate measures of the desirability or plausibility of a state or action. The details of these models are provided in \prettyref{sec:methods}. In brief, they include \textbf{\textit{(i)}} several geometric heuristics (\hyperref[exp:1]{Experiment~1} and \hyperref[exp:2]{Experiment~2}), which measure the distance between the initial and queried object configurations in different ways, and \textbf{\textit{(ii)}} a ``stability'' heuristic (\hyperref[exp:3]{Experiment~3}) which uses a probabilistic stability estimate for each object in the reward (final) configuration \cite{battaglia2013simulation} as a covariate of behavior.

We also evaluated a powerful VLM, OpenAI's ChatGPT v5.0 \cite{chatgpt5}, capable of leveraging statistical regularities learned from large-scale image–text corpora and believed to encode a broad knowledge base \cite{yildirim2024task,huh2024position}. As such, this model serves as a complementary alternative to GenEA's structured, causal agent representation, allowing us to assess whether such an agent representation is necessary. See \prettyref{sec:methods} for further details of each of these alternative models.

\begin{figure}[t!]
\centering
\includegraphics[width=\linewidth]{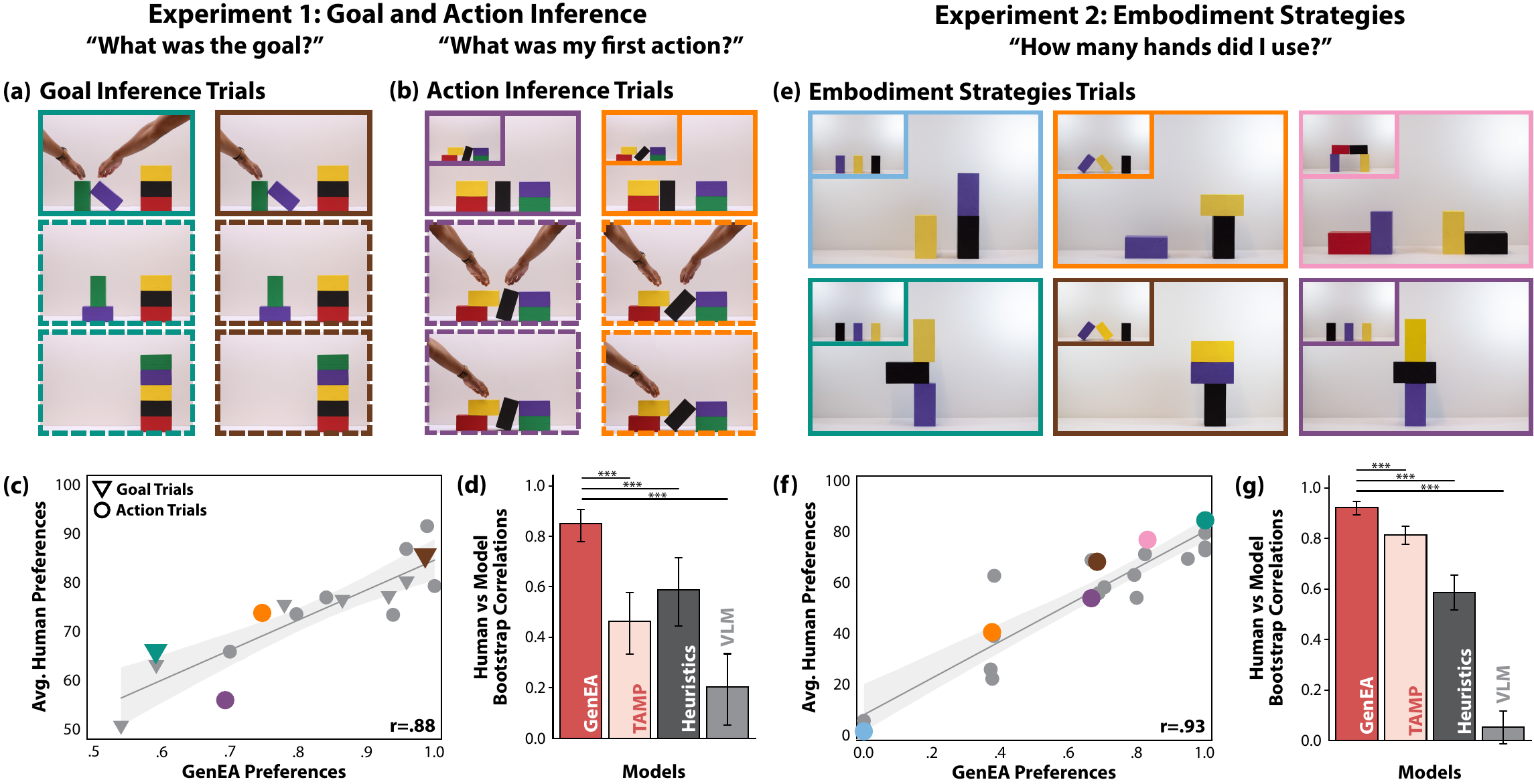}
\caption{ \textbf{Explaining social inferences.}
\hyperref[exp:1]{Experiment 1} comprised two kinds of trials: goal inference and action inference trials. \textbf{(a)} Example stimuli used in the goal inference trials. In these trials, participants viewed an action performed on the initial block configuration and were asked to infer the agent's intended goal from two possible alternatives.   
\textbf{(b)} Example stimuli used in the action inference trials. In these trials, participants were shown the initial and final block configurations and were asked to judge which of two candidate actions produced the given final configuration.  
\textbf{(c)} GenEA strongly correlates with human ratings, reported as the average preference for the option favored by the majority (thus, the axes span half of the scale). The colored data points in the plot correspond to the trials shown in panels (a) and (b), matching their respective frame colors. Note that the y-axis of this scatter plot begins at 50 because any preference for one option logically entails an equivalent lack thereof for the other.
\textbf{(d)} GenEA achieves a significantly stronger correlation with participant judgments compared to the alternatives, including TAMP and heuristics models. Error bars indicate bootstrapped $95\%$ confidence intervals; ``***'' for $p < .001$.
\textbf{(e)} Example stimuli used in \hyperref[exp:2]{Experiment 2} (initial configurations shown in insets), where participants were presented with initial and final block configuration pairs, and asked to report whether the agent who reconfigured the blocks used one hand or two hands.
\textbf{(f)} GenEA closely captures human intuitions in the embodiment strategies task. The colored data points in the plot correspond to the trials shown in panel (e).
\textbf{(g)} GenEA achieves a significantly stronger correlation with participant judgments compared to alternatives, including TAMP and a heuristic model. Error bars indicate bootstrapped $95\%$ confidence intervals; ``***''for $p < .001$.
}
\begingroup
\phantomsubcaption
\label{fig:socialA}
\phantomsubcaption
\label{fig:socialB}
\phantomsubcaption
\label{fig:socialC}
\phantomsubcaption
\label{fig:socialD}
\phantomsubcaption
\label{fig:socialE}
\phantomsubcaption
\label{fig:socialF}
\phantomsubcaption
\label{fig:socialG}
\endgroup
\label{fig:social}
\end{figure}

\section*{Experiment 1: Explaining humans' goal and action inferences}
\label{exp:1} 

Goal and action inference has been a standard paradigm in psychology research for studying human social reasoning \cite{baker2009action,jara2020naive}. We advanced this paradigm to naturalistic, embodied settings with a preregistered experiment \cite{shah2022} examining how people reason about others' plans in manipulation tasks involving building block structures. Participants ($N=60$) completed $18$ goal inference trials ($9$ unique and $9$ color-permuted replicas) and $18$ action inference trials ($8$ unique, $8$ color-permuted replicas plus a third color-permuted replica for $2$ trials, see \prettyref{sec:methods} for details), where the replicas preserved the geometric configuration but altered the blocks' colors. In the goal inference trials, participants observed an agent's action on the initial configuration and were asked to infer which of two final configurations the agent intended to achieve. Therefore, the query for these trials consisted of two reward configurations $Q_R = \{\phi^{(\text{final})}_1,\, \phi^{(\text{final})}_2\}$, one corresponding to each of the trial's two final configurations, and a single plan constraint $Q_{\pi} = {\xi^{(\text{constr})}_1}$ representing the initial conditions presented in the trial. In the action inference trials, participants observed an initial and final configuration pair, and were asked to judge which of two initial actions the agent took to achieve the final configuration. The query for these trials included two plan constraints $Q_{\pi} = \{\xi^{(\text{constr})}_1,\, \xi^{(\text{constr})}_2\}$, each pairing the initial block configuration with one of the two candidate first actions shown in the trial; and $Q_R = \{\phi^{(\text{final})}\}$.

We found that GenEA recapitulates human goal and action inferences, achieving a high degree of alignment ($r=.88$, $p<.001$), significantly stronger than that of any alternative account (\prettyref{fig:socialD}). The correlation plot in \prettyref{fig:socialC} illustrates this relationship, plotting GenEA's preference against humans' for each trial. Points near the lower-left corner of the plot reflect ambivalence, whereas points farther from the origin reflect increasingly strong preference for one of the two options. This visualization shows that GenEA and humans agree not only in terms of the magnitude of their preference (i.e., how much they prefer one option over the other), but also in the direction, consistently favoring the same option.

\section*{Experiment 2: Explaining humans' embodiment strategies}
\label{exp:2} 

Employing such naturalistic, embodied settings affords further leverage on the kinds of investigations we can undertake. Capitalizing on this opportunity, we investigated finer-grained aspects of human action understanding, going beyond asking \textit{what} action would be taken to asking \textit{how} it would be executed (i.e., their embodiment strategies). In a second preregistered experiment \cite{shah2023}, we designed $22$ pairs of initial and final configurations and tasked participants ($N=60$) to report, via a slider bar, their relative preference for using just one hand versus both hands to rearrange each initial configuration into its corresponding final configuration.

As in the previous experiment, we calculated model preferences by querying GenEA (\prettyref{eq:eq2}) with two plan constraints and a single final configuration: $Q_{\pi} = \{{\xi^{(\text{constr})}_1, \, \xi^{(\text{constr})}_2}\}$, where $\xi^{(\text{constr})}_1$ constrains the plans to involve only one-hand actions, and $\xi^{(\text{constr})}_2$ represents plans involving two-hand actions; and $Q_R = \{\phi^{(\text{final})}\}$. In some of our trials, a one-handed solution was not feasible in the model (e.g., the ``green'' trial in \prettyref{fig:socialE}). In these cases, the model assigned a value of $1$, indicating maximum preference for the two-handed option.

Once again, we found that GenEA closely tracked human preferences ($r=.93$, $p<.001$) while outperforming alternative accounts (\prettyref{fig:socialG}). As is evident in the correlation plot (\prettyref{fig:socialF}), GenEA correctly captures not only the clearest cases, where humans preferred either only one-hand (lower-left) or only two-hand (upper-right) options, but also the more graded preferences in between. 

\begin{figure}[t!]
\centering
\includegraphics[width=\linewidth]{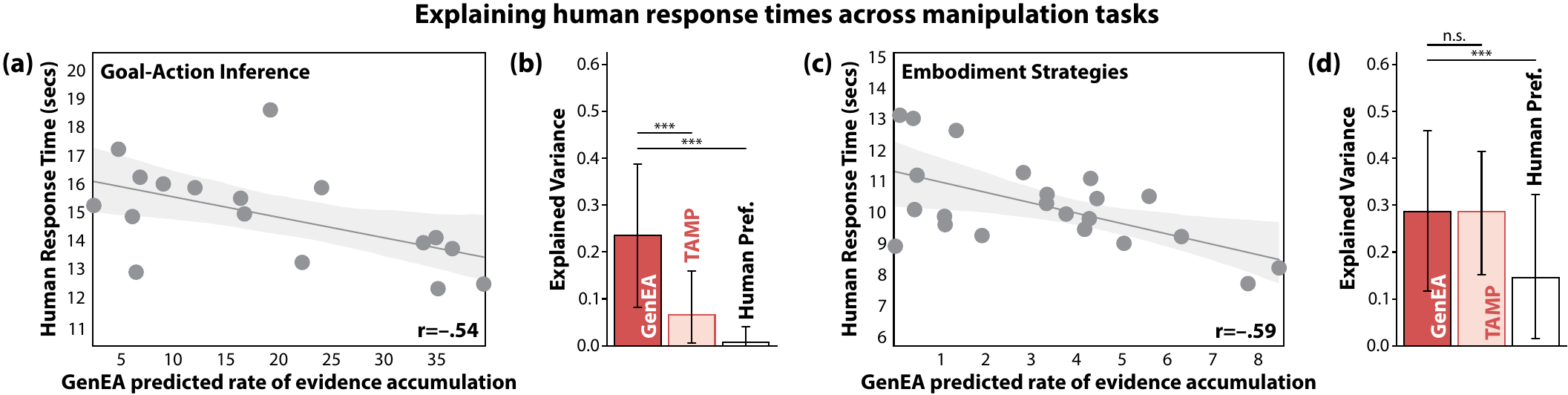}
\caption{\textbf{Explaining human response times.} \textbf{(a)} A scatter plot of GenEA's correlation with human response times in the action and goal inference trials from \hyperref[exp:1]{Experiment 1}. The greater the rate of evidence accumulation predicted by GenEA, the shorter people took to respond. \textbf{(b)} GenEA explains significantly more variance than what TAMP and humans' preference ratings can explain in response times. \textbf{(c)} A scatter plot of GenEA's correlation with human response times in the embodiment strategy inference trials from \hyperref[exp:2]{Experiment 2}. The greater the rate of evidence accumulation predicted by GenEA, the shorter people take to respond.  \textbf{(d)} GenEA explains a similar amount of variance as TAMP and more than what can be explained using participants' preference ratings.}
\begingroup
\phantomsubcaption
\label{fig:thinkA}
\phantomsubcaption
\label{fig:thinkB}
\phantomsubcaption
\label{fig:thinkC}
\phantomsubcaption
\label{fig:thinkD}
\endgroup
\label{fig:think}
\end{figure}

\section*{Explaining humans' response times across manipulation tasks}
The previous section demonstrated GenEA's capacity to account for human preference ratings across two experiments and three manipulation tasks. Alongside these ratings, unbeknownst to the participants, we also logged their response times --- the duration (in seconds) between stimulus onset and response submission. While preference ratings captured the outcome of participants' judgments, this metric offers a complementary source of information about the cognitive processes underlying the assessment and weighing of the two candidate options available in each trial. Response times could in turn be understood as an auxiliary marker of cognitive processes, whereby faster responses reflect judgments that required less processing to arrive at, while slower responses indicate greater processing. Accordingly, we sought to examine whether GenEA could explain the trial-level variability in human response times across the two experiments.

A well-established account of decision time is the drift-diffusion process \cite{usher2001time}, in which the expected response time is determined by the average rate of evidence accumulation favoring one option over the other. GenEA provides a quantity that approximates what would be the rate of evidence accumulation under such a process model: $\sum \vec{v}_{i,\,j} - \sum \vec{v}_{i',\,j'}$ --- the difference between the total probability mass of the two choices in a given trial [$(i,\,j)$ versus $(i^\prime,\,j^\prime)$].

Remarkably, we found that this simple GenEA-derived measure correlated significantly and substantially with humans' response times in both experiments (goal-action inference: $r=-.54, p=.026$, \prettyref{fig:thinkA}; and embodiment strategy: $r=-.59, p=.004$, \prettyref{fig:thinkC}). These correlations are particularly noteworthy given that response times are inherently noisy and were collected during a slider-bar rating task that imposed no constraints on how quickly participants had to respond. 
Furthermore, the bar plots in \prettyref{fig:thinkB} and \prettyref{fig:thinkD} illustrate how GenEA explains as much, if not more, variance than TAMP (GenEA vs. TAMP, goal-action inference $p < .001$ and embodiment strategies $p = 0.94$, using direct bootstrap hypothesis testing), indicating that GenEA's dynamics level (incorporating intuitive physics) substantially contributes to explaining human response times. Critically, GenEA explains significantly more variance in human response times than can be explained by humans' preference ratings (GenEA vs. Human Preference Ratings, goal-action inference $p < .001$ and embodiment strategies $p < .001$), indicating that GenEA captures more than a simple relationship between accuracy and response time. 

\begin{figure}[t!]
\centering
\includegraphics[width=.8\linewidth]{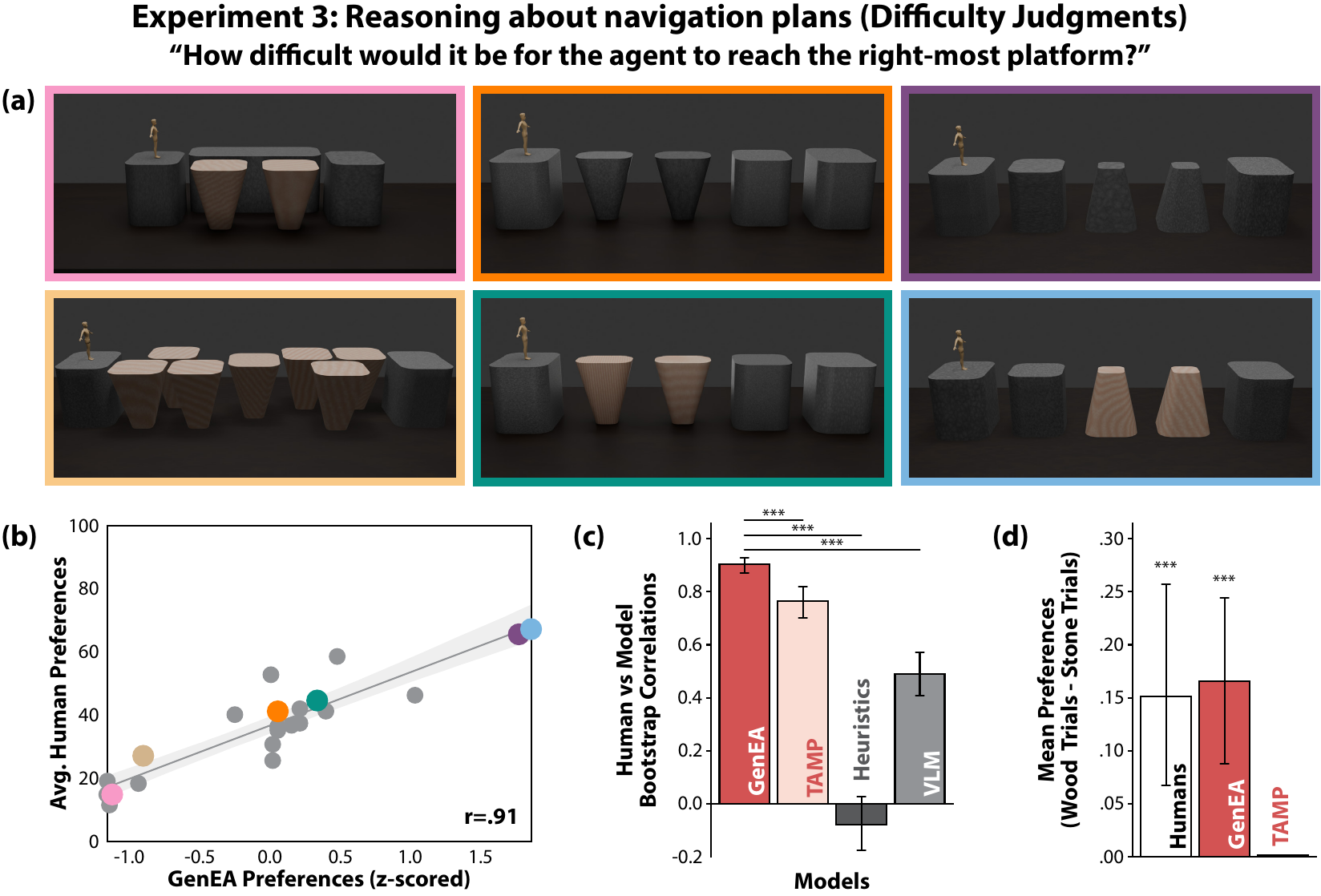}
\caption{\textbf{Testing the generality of GenEA in the domain of navigation.}
\textbf{(a)} Example stimuli from the navigation task of \hyperref[exp:3]{Experiment 3}, showing an agent standing on the left-most ``start'' platform, one or more intermediate platforms and the ``final'' platform to the extreme right. On each trial of this experiment, participants reported their judgment of how difficult it would be for the agent to reach the right-most platform.
\textbf{(b)} GenEA closely captures people's difficulty judgments ($r=.91,\, p < .001$). Colored dots correspond to the data points of the trials visualized in panel (a). 
\textbf{(c)} GenEA explains more variance than alternatives, including TAMP and heuristic (stability) models.
\textbf{(d)} Both GenEA as well as humans, but not TAMP, show sensitivity to the material composition of the intermediate platforms, with both judging trials with wooden platforms to be on average more difficult to traverse than those with stone platforms. }
\begingroup
\phantomsubcaption
\label{fig:diff_navA}
\phantomsubcaption
\label{fig:diff_navB}
\phantomsubcaption
\label{fig:diff_navC}
\phantomsubcaption
\label{fig:diff_navD}
\endgroup
\label{fig:diff_nav}
\end{figure}

\section*{Experiment 3: Generalizing GenEA beyond manipulation to navigation}
\label{exp:3}

Although our ability to reason about embodied plans is not confined to any single domain, computational accounts of physical and social reasoning have nonetheless predominantly been developed and tested within narrow, domain-specific settings, such as social inferences in grid worlds, physical reasoning in block structures \cite{fahlman1974planning,battaglia2013simulation, jaraettinger2024theory,ying2026grounding}. 
Any candidate for the agent representation in the mind must account for human behavior beyond any single domain, like that of reasoning about plans in naturalistic manipulation settings. Therefore, here we evaluate GenEA in the qualitatively distinct domain of navigation. Moreover, as an additional test of the generalizability of our proposal, we consider an entirely different class of tasks, moving from making social inferences about an agent's actions and goals to making judgments about more abstract, higher-order properties of plans. Whereas goals and actions are constitutive of a plan itself --- a plan is defined by the goal it serves and the actions that carry it out --- higher-order properties are instead mere readouts over the plan, raising the question of what might underlie such judgments, particularly when they are consistent across individuals. We sought to characterize one such property: judgments of task difficulty \cite{gweon2017reverse}.

In a new preregistered study \cite{shah2026}, participants ($N=60$) evaluated the difficulty of navigating obstacle-course-like configurations, each comprising a unique sequence of platforms leading from a designated ``start'' platform (always presented on the left) to a ``final'' platform (always presented on the right). Across the $22$ trials we designed, the intermediate platforms varied either in shape (for instance, perfectly cuboidal or tapered toward the base) or in their material composition, either wood or stone, intended to convey differences in mass. Although participants were not explicitly provided with information about how to interpret these material properties, they were instructed to assume that real-world physics applied, including concepts such as gravity and that an object's mass depends on its material composition. On each trial, an agent was shown standing on the ``start'' platform, and participants were tasked with using a slidebar to respond to the prompt: ``How difficult would it be for the agent to reach the right-most platform?''

Given the information available to participants, we limited the GenEA query to the initial and final object configurations: $Q_R=\{\phi^{(\text{final})}\}$ and $Q_{\pi}=\emptyset$. Note that, unlike the two experiments described above, this experiment did not present participants with options, but instead required that they report a single difficulty rating on each trial. To model these difficulty ratings, we first identified the maximum of the queried distribution under GenEA (\prettyref{eq:eq2}), $\max \vec{v}_{i,\,j}$), corresponding to the most probable plan given the reward, and used the Negative Log-Likelihood score of this plan (equivalent to its cost) as the model's prediction for each trial. For both GenEA and TAMP, we generated $60$ runs per trial, matching the number of human participants, so as to obtain a stable estimate of the measure, in the presence of uncertainty arising from the noisy kinematic trajectories and the application of dynamics.

GenEA closely recapitulated human difficulty judgments ($r=.91,\,p<.001$; \prettyref{fig:diff_navB}), while significantly and substantially outperforming alternatives (\prettyref{fig:diff_navC}). Overall, both humans and GenEA consistently judged configurations that can be traversed by walking as easier than those requiring jumps. Importantly, unbeknownst to the participants, we had deliberately embedded two design features into the stimuli set to probe auxiliary aspects of human reasoning about embodied plans. First, we included configurations that admitted multiple candidate paths. The pink trial in \prettyref{fig:diff_navA}, for example, contains a single long, stable platform affording an easy, walkable route. Both humans and GenEA ground their difficulty judgments in this route rather than in the two comparatively unstable platforms in the foreground, and accordingly judge the pink trial among the easiest. Second, we included seven matched pairs in the stimulus set, differing only in the material composition of some or all intermediate platforms, allowing us to isolate the effect of platform material on difficulty judgments, insofar as lower-mass platforms would be expected to be less stable and more prone to toppling. For each human participant and each model run, we computed the difference between the averaged difficulty judgments across the wood and stone trials. \prettyref{fig:diff_navD} presents the resulting bootstrapped mean differences. Human participants and GenEA show comparable sensitivity to this manipulation, with the direction of the effect also remaining consistent and statistically significant across both: trials with wooden platforms are judged, on average, to be more difficult to traverse. TAMP, by contrast, shows no sensitivity to the change in the platform's material composition, underscoring the importance of incorporating intuitive physics (the dynamics level) into the model.

Finally, to offer deeper qualitative insight into GenEA's ability to explain human inferences across these three experiments, we discuss a subset of trials for each experiment in the Supplement (\prettyref{suppfig:supp_social_goal}, \prettyref{suppfig:supp_social_action}, \prettyref{suppfig:supp_hands}, and \prettyref{suppfig:supp_diff}), including those marked by color in the scatter plots.

\section*{Discussion}

We presented GenEA, a novel agent representation that characterizes and computationally explains how humans reason about plans in natural, embodied settings. In \hyperref[exp:1]{Experiment~1} and \hyperref[exp:2]{2}, we showed that GenEA captures humans' goal and action inferences as well as their judgments about embodiment strategies (such as whether an agent used one or two hands) in object manipulation tasks. GenEA also explains participants' response times in these two experiments, indicating that GenEA not only captures the end products of human reasoning but also important aspects of the underlying cognitive process. In \hyperref[exp:3]{Experiment~3}, we demonstrated that GenEA generalizes to the qualitatively distinct domain of navigation, successfully accounting for a higher-order property of plans, namely human judgments of task difficulty. Across all three experiments, GenEA consistently achieved strong quantitative fits and captured key patterns in human behavior --- including sensitivity to subtle variations in object configuration or material composition built into the stimulus set (see \prettyref{suppfig:supp_social_goal}, \prettyref{suppfig:supp_social_action}, \prettyref{suppfig:supp_hands}, and \prettyref{suppfig:supp_diff} for details) --- while also outperforming a powerful set of alternative models.

The fields of brain and cognitive sciences have made remarkable strides by focusing on distinct domains of cognition \cite{kanwisher2010functional,spelke2024precis} and investigating separate knowledge systems that support these areas. These advances pose a foundational and pressing question: How do these disparately studied parts of cognition --- such as object perception \cite{dicarlo2007untangling,bi2026visual}, social cognition \cite{jara2016naive,jara2020naive,jaraettinger2024theory}, and intuitive physics \cite{kubricht2017intuitive} --- come together to form a cohesive cognitive architecture that enables flexible reasoning and behavior? Our work, which lies at the intersection of multiple cognitive domains, serves as a case study for productively exploring this question. GenEA motivates a new perspective on the architecture of cognition, in which the agent representation is factorized into its goals, its body's coarse biomechanics, and how its actions cause and exploit object motion. GenEA suggests that our versatility in reasoning about embodied plans stems from connecting an agent's abstract goals to its bodily actions and the physical environment via a multilevel state space. In this way, GenEA reveals how agent-based intuitive psychology is deeply intertwined with object-based intuitive physics --- two domains that have largely been studied independently of one another \cite{liu2024naive}. 

Within Marr's levels-of-analysis framework \cite{marrvision}, we regard GenEA as a computational-level account of how people reason about embodied plans. That is, we argue that our work identifies the problem being solved when people draw such inferences, namely computing joint and conditional distributions under a generative model structured by a multilevel agent representation resembling GenEA, rather than specifying the particular algorithm by which that problem is solved. It is possible that, in the mind, GenEA's posterior distribution is approximated via a collection of smaller forward models and modules \cite{wolpert1994forward,jacobs1999computational}, which together may instantiate some multilevel agent representation. Given the evidence we presented in this paper, our claim is that any such representation must be broadly analogous to the one implemented in GenEA. Lending further credence to this claim is the fact that several individual components of GenEA, including symbolic, tree-based planning \cite{van2023expertise}, simulation-based intuitive physics \cite{battaglia2013simulation}, and posture-based planning \cite{rosenbaum2001posture}, have each independently received empirical and computational support in the literature. Additional converging evidence comes from the finding that GenEA's parametrization based on kinetic energy (\prettyref{eq:eq2}) outperforms less embodied alternatives, such as plan length \prettyref{suppfig:supp_temp_prob_pearson}, positioning our modeling framework as a promising starting point for future algorithmic explorations of reasoning about embodied plans.

Having validated GenEA across a diverse set of tasks (social inference, embodiment strategies, difficulty judgments) that span navigation and object manipulation, we suggest that the same framework can plausibly be extended to reason about other higher-order properties of agents, such as skill or competence \cite{zhang2024people,leonard2019better,muradoglu2020children,xiang2023collaborative}, motivation \cite{chong2017neurocomputational}, and persistence \cite{leonard2017infants}.
A similar architectural pattern may underlie other ``interfaces'' of cognition. Within affective cognition, for instance, empathy when cast as inferring others' hidden affective state, like physical pain \cite{berke2026toward}, may be formalized in terms of a generative model embedding GenEA-like factorized representations of body and the physical context. 
Furthermore, GenEA is not restricted to inferences about a ‘generic other’; rather, its agent representation, $\beta \in \xi$, can be adaptively tailored to a specific individual, readily supporting idiosyncratic inferences about familiar individuals, including one’s own self \cite{paul2026reverse, brecht2017body}, such as individual-specific goals, actions, physical capabilities, and physical limitations. 
GenEA could also support drawing rich social inferences from mere physical ``traces'' left behind by the embodied actions of others \cite{lopez2021detecting,lopez2022social,jara2024traces,wu2024whodunnit}. 
Beyond its central focus on reasoning about embodied plans, GenEA also bears on the human planning process itself. The planning algorithm we implemented over GenEA's multilevel state-space constitutes, in its own right, a novel extension of TAMP that incorporates intuitive physics. So, GenEA could, in principle, explain how human participants would actually build block structures, for instance, by detailing their action sequences, or estimating the time spent deliberating over decisions and executing those actions \cite{nuzzi2026continuous}. 

Finally, we believe that GenEA, when extended with perception systems that make it image-computable, could inform the development of robotics and artificial intelligence systems capable of making rich social inferences \cite{gweon2023socially,jara2019theory} in complex naturalistic environments. For robots to leave the laboratory and seamlessly interact with humans in everyday physical settings --- whether as collaborators in physical tasks, household assistants, or social companions --- it is crucial that they understand and can reason about embodied actions of humans \cite{kelley2010understanding}. In this context, GenEA provides a proof-of-concept system with cognitive validity that can complement existing work focusing on creating large datasets and learning predictive models from them (e.g., \citenum{thompson2024predicting,hoffman2024inferring}). We suggest that these systems, much like GenEA, should incorporate rich generative models of embodied action planning. 

\section*{Materials and Methods} 
\label{sec:methods}

\subsection*{Explaining human inferences across distinct domains}
Our experiments were conducted across two distinct domains: manipulation and navigation. In the manipulation setting, we considered a setup in which humans and the computational models reasoned about the goals and embodied plans involved in reconfiguring an initial configuration of objects (blocks) into a given final configuration. In the navigation setting, humans and the computational models rated the difficulty of traversing a sequence of platforms leading from a designated start platform to a given final platform. 

We evaluated human performance against that of the models in three reasoning tasks: goal and action inference (\hyperref[exp:1]{Experiment~1}), inferences about embodiment strategies such as whether to use one or two hands (\hyperref[exp:2]{Experiment~2}), and judgments of task difficulty (\hyperref[exp:3]{Experiment~3}). In each reasoning task, humans and the models were provided with the initial object configuration, one or more final (goal) configurations, and, where applicable, one or more constraints on the actions available to the agent from the initial object configuration. 

\subsection*{GenEA's multilevel state-space, and the multilevel planning within it}
GenEA is based on a multilevel agent representation $\xi = \{\beta,\, \phi,\,f^{(\text{S})},\,f^{(\text{K})},\,f^{(\text{D})}\}$, characterizing the coupled interactions between the coarse biomechanics of the body $\beta$ and the dynamics of $N$ nearby objects (blocks or platforms) $\phi$, positioned atop a rigid, immovable supporting surface. Changes to the state of $\beta$ or $\phi$ are governed by the functions $f^{(\text{S})}$, $f^{(\text{K})}$ and $f^{(\text{D})}$. The multilevel planning process searches over the resulting state space, sequentially applying symbolic to kinematics to dynamics level functions. \prettyref{alg:alg1} presents this planner, and the following sections provide an in-depth description of each of its components.

\begin{algorithm*}
    \caption{The GenEA planner}
    \begin{algorithmic}[1]
        \Inputs{$\xi^{(\text{init})},\xi^{(\text{final})}$}
        
        \Initialize{$\xi^{(\text{root})} = \xi^{(\text{init})} ;\; Q = \Call{CreateQueue}{}();\; \Call{Enqueue}{Q,\, \xi^{(\text{root})}}$}
        \Repeat
            \State $\xi^{(\text{current})} \leftarrow \Call{Dequeue}{Q}$
            \If {$\Call{StoppingCondition}{\xi^{(\text{current})},\, \xi^{(\text{final})}}$}
                \State $\Call{SetIsValid}{\xi^{(\text{current})},\, True}$; $\textbf{continue}$
            \EndIf
            \State $\text{A}^{(\text{S})} \leftarrow \Call{SymbolicAutomaton}{\xi^{(\text{current})}}$
            \If {$A^{(\text{S})} \, \textbf{is not} \, \emptyset$}
                \State $\Call{SetIsValid}{\xi^{(\text{current})},\, True}$
                \For {\textbf{each} $\text{a}^{(\text{S})}$ in $\text{A}^{(\text{S})}$}
                    \State $\xi^{(\text{K})},\, \tau^{(\text{K})} \leftarrow \Call{KinematicSolver}{\xi^{(\text{current})},\, \text{a}^{(\text{S})}}$  \Comment{determine and execute action trajectories}
                    \State $\xi^{(\text{D})} \leftarrow \Call{ForwardEngine}{\xi^{(\text{K})}}$  \Comment{run physics-based simulation}
                    \State $C \leftarrow \Call{ComputeCost}{\xi^{(\text{current})},\, \tau^{(\text{K})}}$ 
                    \State $\Call{Enqueue}{Q,\, \xi^{(\text{D})}}$  \Comment{queue $\xi^{(\text{D})}$ for future explorations}
                    \State $\Call{AddChild}{\xi^{(\text{current})},\, \xi^{(\text{D})}}$  \Comment{update the search tree}
                    \State $\Call{SaveData}{\xi^{(\text{current})},\, C,\, \text{a}^{(\text{S})}}$
                \EndFor
            \EndIf
        \Until{$Q \, \textbf{is} \,  \emptyset$}  \Comment{until there is nothing left to explore}
    \end{algorithmic}
    \label{alg:alg1}
\end{algorithm*}

\subsubsection*{Implementing the multilevel state-space $\xi$ for manipulation tasks}

For block building tasks, we implement $\xi$ by extending the ``Configuration'' data structure of  \textproc{KOMO} ($k^{th}$-order Markov Optimization) \cite{toussaint2015logic}. This data structure describes an indexable array of 3D shapes (dimensions of each block and each body part), as well as the kinematic (position, orientation, velocity) and dynamical properties (mass, friction) of the agent's body parts and each block. In all our stimuli, each block is either a right rectangular prism ($.05$ units in width and length, with a height of .1 units) or a cube (each side measuring .05 units). The blocks are arranged in a block-tower-like configuration on a stationary rectangular tabletop. A stimulus can feature multiple blocks with the same shape and color.

The ``Configuration'' data structure also includes another linked data structure called ``joints'', which specifies the joint actuators of the humanoid body in a very traditional sense. ``Joints'' are also created and deleted if an actuator grasps (via \textproc{Pick}) or releases an object (via \textproc{Cache} or \textproc{Fix}). In a similar regard, ``joints'' also enables the construction of a directed acyclic ``support graph'' to capture the arrangement of blocks on the tabletop. In this graph, the nodes represent blocks, with the tabletop serving as the root node, and edges indicate the support relations between two blocks. The acyclic nature of the graph mandates that each node can only be supported by one other node. So, for instance, if a block rests on top of two other blocks, the support graph will create an edge connecting it to only one of the two underlying blocks, with no particular preference for either.

\subsubsection*{Implementing the multilevel state-space $\xi$ for the navigation task}

For navigation tasks, we implement $\xi$ in \textproc{Crocoddyl} \cite{mastalli2020crocoddyl} using data structures provided by \textproc{Pinocchio} \cite{carpentier2019pinocchio} to represent the agent $\beta$ and its kinematic properties, and the data structures from \textproc{Mujoco} \cite{todorov2012mujoco} to represent the kinematic (position, orientation, velocity, etc.) and dynamic properties (mass, friction, etc.) of the platforms. For the computational model, we represented the agent using the Talos humanoid robot \cite{stasse2017talos}, including only the joints from the hips downward. \textproc{Crocoddyl}, an efficient differential dynamic programming solver, provided the framework for implementing the associated symbolic actions: walking (via \textproc{Step}), jumping (via \textproc{Jump}), and turning (via \textproc{Turn}).

\subsubsection*{Planning within the multilevel state-space $\xi$}

The multilevel planning algorithm takes as input an initial state $\xi^{(\text{init})}$ and a final state $\xi^{(\text{final})}$. The initial state consists of $\phi^{(\text{init})}$, the initial object configuration, and $\beta^{(\text{init})}$, the coarse humanoid body in a standard upright posture (as shown in \prettyref{fig:computational_modelA}, top). The final state consists of $\phi^{(\text{final})}$, the final object configuration, and $\beta^{(\text{final})}$, the configuration the humanoid body assumes after the last action of a plan. 

\subsubsection*{Task planning with $f^{(\text{S})}$}

The task planner identifies high-level symbolic actions, $\text{a}^{(\text{S})} \in \text{A}^\text{S}$, that, when applied to a given initial object configuration, bring it closer to the final object configuration. This is achieved through a finite state automaton that governs which symbolic actions are available to the agent based on the state of the agent $\beta$ and objects $\phi$, and is denoted as $f^{(\text{S})}: \{\beta,\, \phi\} \rightarrow \text{A}^{(\text{S})}$. Using this automaton, the task planner defines a search tree that maintains several ordered lists of states $\xi$, each beginning with the initial state $\xi^{(\text{init})}$  and ending with the final state $\xi^{(\text{final})}$. The edges connecting consecutive nodes of this search tree are the symbolic actions of the agent. They are symbolic because they do not describe how these actions can actually be implemented by moving the agent's body (or when applicable, the objects the agent holds). Instead, they provide ``targets'' for motion planning, which ultimately result in discrete intermediate $\xi$s that become part of the search tree. 

We perform a breadth-first traversal to navigate and expand this search tree, which uses the support graph of the final object configuration $\phi^{(\text{final})}$ to determine which object or objects should be manipulated or moved to at each step. 
In expanding the tree, the rules associated with each of the symbolic actions are listed as follows. 

Rules for \textproc{Fix}: \hskip2mm 
\begin{enumerate*}[(i),font=\bfseries,itemjoin={\hskip2mm}]
\item The block's final location, as defined in $\xi^{(\text{final})}$, must be free and available. 
\item A block can only be ``Fixed'' at locations that are themselves already fixed (i.e., the supporting blocks must be in their respective final positions). For instance, when fixing a block onto a stack of blocks, all the blocks in that stack must have been fixed already. The tabletop, which cannot be moved, is always considered fixed.
\end{enumerate*} 

Rules for \textproc{Cache}: \hskip2mm
\begin{enumerate*}[(i),font=\bfseries,itemjoin={\hskip2mm}]
\item A block can only be cached at one of two predesignated locations on the tabletop. 
\item These locations must be available, i.e., not already occupied by another block. Each cache location can only accommodate one block. 
\item When determining the cache location, use the nearest (Euclidean distance) of the two possibilities. 
\item Block that can be ``Fixed'' cannot be ``Cached''. 
\end{enumerate*}

Rules for \textproc{Pick}: \hskip2mm
\begin{enumerate*}[(i),font=\bfseries,itemjoin={\hskip2mm}]
\item Only one block can be picked at a time by a single hand.
\item A block cannot be picked if it was just placed at the previous time step. 
\item The algorithm decides which hand to use for the \textproc{Pick} operation based on the hands' proximity (Euclidean distance) to the target block. If both hands are equidistant from the block, the hand assignment is done based on the block's intended final position. For instance, if the block is to be placed on the right half of the work desk, the right hand is selected, and vice versa. 
\item At most two (different) blocks can be picked simultaneously, one with each hand. 
\item We assume a small margin of error around the final position, defined as the sphere of radius $.018$ centered at the final position. If an object's center lies outside of this radius, it is not marked as ``Fixed'' and, therefore, is available to be ``Picked''. 
\end{enumerate*}

In navigation settings, the agent explores all available platforms within a search radius of $1.0$ units at each state, with the action possibilities determined by the distance between its current location and the nearest edge of a candidate platform. Two rules determine which actions are admissible: the agent may not \textproc{Jump} when walking (via \textproc{Step}) is possible, and it may neither \textproc{Step} onto nor \textproc{Jump} to a platform that it has already visited.

If the pruning rules described above indicate that no actions can be taken at a node, no further exploration occurs on that branch.

\subsubsection*{Motion planning with $f^{(\text{K})}$}
Given a symbolic action, the motion planner is invoked to generate the trajectories, $\bm{\tau}^{(\text{K})}$, updating the state-space with additive noise, $\{\beta,\,\phi\} \xrightarrow{\bm{\tau}^{(\text{K})} + \epsilon} \{\beta^{(\text{K})},\,\phi^{(\text{K})}\}$. 
We use a truncated skewed normal distribution to model this additive noise, which provides sufficient flexibility to accommodate various symbolic actions. Importantly, these parameter values were not fit to human behavior but were instead chosen to satisfy the geometric constraints associated with each action: 
\begin{enumerate*}[(i),font=\bfseries,itemjoin={\hskip2mm}]
\item \textproc{Fix} introduces a small amount of perturbation to the endpoints of its trajectories, modeling a naive form of motor noise, via a truncated skewed normal distribution centered at $0$, with standard deviation $\sigma = 0.01$ and skewness parameter $\alpha = 0$. The perturbations are bounded to the interval $[-0.01,0.01]$ units to avoid unrealistic releases that could artificially induce instability.
\item \textproc{Step} introduces endpoint perturbations to model uncertainty in the agent's final position at the end of the walk. Perturbations are sampled from a truncated skew-normal distribution centered at $0$, with standard deviation $\sigma=0.1$ and skewness parameter $\alpha=-1$. The negative skew increases the probability that the agent will stop closer to the intended target location. 
\item \textproc{Jump} introduces endpoint perturbations to model uncertainty in the agent's landing position. Along the direction of the jump, perturbations are sampled from a truncated skew-normal distribution centered at $0$, with standard deviation $\sigma=0.5$ and skewness parameter $\alpha=-1$, increasing the probability of landing closer to the intended target location. Perpendicular to the direction of the jump, perturbations are sampled from the same truncated skew-normal distribution with $\sigma=0.5$ and $\alpha=0$, yielding symmetric uncertainty around the intended landing position. These perturbations are bounded to the interval $[-1.25, 1.25]$ to prevent unrealistic jumps.
\end{enumerate*} 

In manipulation settings, trajectory optimization is implemented using \textproc{KOMO}, which specifies a mathematical program (i.e., a set of inequality and equality constraints) that, when solved (via optimization), yields these trajectories for the given symbolic action. It takes as input a symbolic action $\text{a}^{(\text{S})}$ and $\xi_t$ to populate the set of inequality and equality constraints, and solves it to determine the position and orientation that the end-effector must assume at each time-discretized waypoint (per symbolic action, the trajectories are defined over $200$ time steps). It then performs a kinematic inversion to convert these end-effector poses into joint values, producing an executable trajectory for the agent. This process is denoted as $f^{(\text{K})}: \{\beta,\, \phi,\, \text{a}^{(\text{S})}\} \rightarrow \bm{\tau^{(\text{K})}}$ in the main text; readers interested in further details of this trajectory optimization framework should consult \cite{toussaint2014newton}. GenEA and the TAMP-based model extend an open-source implementation of this framework; see \cite{toussaintRai2024}. The motion planner executes this trajectory to produce an intermediate $\xi^{(\text{K})}$.

In navigation settings, trajectory optimization is implemented using \textproc{Crocoddyl}, which formulates and solves a differential optimal control problem incorporating the agent's forward dynamics and contacts. Given a symbolic action $\text{a}^{(\text{S})}$ and $\xi_t$, \textproc{Crocoddyl} discretizes the trajectory using an integration step of $0.02$ seconds. The number of discretization knots varies across action phases. A walking action comprises $4$ knots for the standing phase and $20$ knots for each foot's swing phase, with an additional impulse model capturing foot contact at the end of each swing. A jumping action comprises $20$ knots for take-off, $20$ knots for upward flight, $20$ knots for downward flight, and $10$ knots for landing, with an additional impulse model capturing contact upon touchdown. The resulting trajectory specifies the agent's joint configurations and control inputs required to execute the symbolic action.

\subsubsection*{Intuitive Physics with $f^{(\text{D})}$}
We follow \cite{battaglia2013simulation} in modeling intuitive physics using a game-engine-style physics engine to capture humans' expectations of how objects react to external forces (e.g., gravity, object-object interactions). For manipulation tasks, we use NVIDIA's \textproc{PhysX} engine \cite{nvidiaPhysX}, and for navigation tasks we use DeepMind's \textproc{MuJoCo} engine \cite{todorov2012mujoco}. In both cases, gravity is modeled as a uniform downward acceleration of $9.8\,\mathrm{m/s^2}$, with simulations running for $2$ seconds in manipulation and $5$ seconds in navigation. For \textproc{PhysX}, we use the engine's default values for dynamical properties, including mass and friction. In \textproc{MuJoCo}, we assign wooden platforms a mass of $1$ unit and stone platforms a mass of $9$ units with frictionless contact.

The simulations update the multilevel state-space via $f^{(\text{D})}: \phi^{(\text{K})} \rightarrow \phi^{(\text{D})}$, resulting in $\xi^{(\text{D})}$, which serves as input to the symbolic task planner in the next planning step.

\subsection*{Joint distribution over object manipulation goals and embodied plans}

The key computational substrate of GenEA is the joint distribution over rewards $R$ and embodied plans $\pi$, identified in \prettyref{eq:eq1} in the main text. This joint distribution is factorized into a uniform prior distribution over goals Pr$(R\,|\,\xi)$ and a conditional distribution over plans given a reward (final) configuration, Pr$(\pi\,|\,R,\,\xi)$.  We construct this latter conditional distribution using all the plans returned by the planner given a reward, with the probability of each plan inversely proportional to its cost. 

The results reported in the main paper used the kinetic energy of a plan as the cost function, calculated from the trajectories of end-effectors specified in the respective symbolic actions. See next section for alternative formulations of the cost function and the robustness of our results to that choice.

As stated in the main text, reasoning queries (\prettyref{eq:eq2}) always contain at least one reward, e.g., $\phi^{(\text{final})}$. This leads to the query of Pr$(\pi,\,R\,|\,\xi)\cdot\delta_{R=\phi^{(\text{final})}}$, which is equivalent to the conditional Pr$(\pi\,|\,R,\, \xi)$. Thus, in practice, we operate with this conditional, to which we can further factor in plan constraints (e.g., $\xi^{(\text{constr})}$), if it is included in the query.

\subsection*{Robustness of our results to alternative formulations of the cost function and the temperature parameter}

Even though the kinetic energy is a reasonable choice for the cost function to reason about embodied plans, there is no a priori reason that it's the cost function humans consider. To make sure that our results are not sensitive to this particular choice of cost function, we conducted additional analysis using two other possibilities: the symbolic plan length, derived from the search tree, and the ``solver-derived optimization costs'', namely either ``sum of squared accelerations'' obtained directly from the \textproc{KOMO} package \cite{toussaint2014newton} for manipulation tasks, or ``sum of squared residuals'' obtained directly from the \textproc{Crocoddyl} package \cite{mastalli2020crocoddyl} for tasks involving navigation,.

\prettyref{suppfig:supp_cost_funcs} presents bar plots of the bootstrapped correlations between human responses and GenEA, using each of these alternative cost functions, across all experiments. We can see that, irrespective of the choice, GenEA consistently outperforms alternative models. \prettyref{suppfig:supp_temp_prob_pearson} shows that our results are robust to the choice of the temperature parameter $\kappa$. \prettyref{suppfig:supp_temp_prob_pearson} also shows that the kinetic-energy-based parametrization of GenEA leads to more consistent behavioral matches than any other cost functions we tested.

\subsection*{Details of the alternative models compared against GenEA}

The TAMP-based model was identical to GenEA except that its multilevel state-space, denoted $\xi^{(\text{TAMP})}$, did not include the intuitive physics level $f^{(\text{D})}$. We also considered a heuristics model based on the stability of object configurations for \hyperref[exp:3]{Experiment~3}. Following \cite{battaglia2013simulation}, we perturbed each platform by applying forces at random locations on its top surface; and consistent with this setting, the forces were primarily directed downward. The magnitude and direction of the applied force were sampled from skewed normal distributions with mean $-1$, standard deviation $5$, skewness parameter $\alpha=-5$, and bounds of $[-5,5]$. These values were calibrated to match the range of forces produced by \textproc{Crocoddyl}. For each platform, we ran $1000$ independent simulations with these perturbations and counted the number of simulations in which the platform became unstable. We then summed these instability counts across all platforms in a stimulus to obtain the covariate for modeling behavior, with higher values indicating greater instability and, consequently, greater predicted task difficulty. A simulation was classified as unstable if the platform deviated from its original pose (position or orientation) by more than $0.1$ units following the applied perturbation.

In addition to the TAMP model and stability heuristic, we designed additional experiment-specific heuristic models to challenge the core commitments presented in our paper regarding how humans reason about embodied plans. Following our preregistration \cite{shah2022}, in the action inference trials of \hyperref[exp:1]{Experiment~1}, excluding the ``Picked'' blocks, the heuristic model favored the option that involved the fewest remaining blocks to be moved. The preference was measured as the ratio of the remaining blocks to be moved for the selected option to the total number of blocks to be moved across all options. Following our preregistration \cite{shah2022}, in the goal inference trials of \hyperref[exp:1]{Experiment~1}, not considering the ``Picked'' block(s), the heuristic model selected the option with the least Euclidean distance between blocks in their initial configuration and the respective final configurations. The preference was recorded as the ratio of the distance remaining to be covered by the selected option and the total remaining distance across all options. Following our preregistration \cite{shah2023}, in \hyperref[exp:2]{Experiment~2} evaluating human embodiment strategies, the heuristics model assigned a value to each trial depending on the number of blocks moved between the initial and final configurations. It registered a value of ``0'' when the number of blocks moved was 1 and a ``100'' on the slider when it was greater than 1.

We used ChatGPT v5 \cite{chatgpt5}, one of OpenAI's most powerful models, as the Video-Language Model (VLM) of choice. We provided the model with a detailed description of the task and explicit instructions regarding the expected output. In addition, to ensure task comprehension, the VLM was prompted with a few binary verification tasks. These were presented as conversational snippets involving questions to the VLM (user messages) that are paired with the corresponding ground-truth response that the VLM was expected to produce (assistant messages). Only after successfully completing these verification steps was the model prompted to report a preference on any given trial. Please note that no model weights were updated during this process. For details on the VLM prompts, please refer to the supplementary section: \prettyref{suppfig:supp_vlm_exp_action} for action inference trials of \hyperref[exp:1]{Experiment~1}, \prettyref{suppfig:supp_vlm_exp_goal} for goal inference trials of \hyperref[exp:1]{Experiment~1}, \prettyref{suppfig:supp_vlm_exp_hands} for \hyperref[exp:2]{Experiment~2}, and \prettyref{suppfig:supp_vlm_exp_nav_diff} for the difficulty judgment trials of \hyperref[exp:3]{Experiment~3}.

\subsection*{Behavioral Studies}

In our analysis, we show correlations between the averaged participant data and the z-scored model predictions. These correlations were derived from 1000 bootstrap simulations. In each bootstrap simulation, we sampled participants with replacement, calculated the average scores for each trial in this participant pool, and then computed the correlation between these scores and the model's z-scored values.

\subsubsection*{Experiment 1: Explaining humans' goal and action inferences}

This experiment was pre-registered \cite{shah2022}, including the stimuli, procedure, exclusion criteria, number of participants, and analysis.

\textbf{Stimuli:} 
It included two trial types: $8$ unique Goal Inference (Goal) trials (see \prettyref{suppfig:supp_stim_social_goal} for the entire set of stimuli) and $9$ unique Action Inference (Action) trials (see \prettyref{suppfig:supp_stim_social_action} for the entire set of stimuli). For each of these stimuli, we designed an additional replica that preserved the overall block configuration but featured blocks that were differently colored. We also constructed a third color-permuted replica for two Action trials and used it as an ``attention check'' to gauge participant engagement during the experiment. Therefore, a total of $36$ trials were conducted per participant ($9 \times 2$ Action trials, $8 \times 2$ Goal trials, and $2$ attention check Action trials). 

Each stimulus was presented as an image displaying a configuration of five blocks. All blocks had the same shape (height was twice as much as the width) but different colors (red, yellow, blue, purple, and green). Some images additionally included one or two hands of a real human agent, indicating the targeted block(s) for interaction. 

\textbf{Procedure and Participants:}
In the action inference trials, participants were presented with the initial and final block configurations (without any agent) and two initial actions (showing an agent's hand(s) acting on the initial configuration). Their task was to rate which of the two actions they think the agent might have taken first to move the initial configuration to the final configuration. 

In the goal inference trials, participants were shown the initial block configuration, with an action performed on it, and two alternative goal configurations (without any agent). Their task was to rate which of the two final block configurations corresponded to the agent's goal. 

The presentation order of the trials was semi-randomized between participants, ensuring that no more than two consecutive trials belonged to the same trial type (e.g., participants are shown at most two Goal trials in succession). Additionally, since some of the trials contained the same configuration (but a different permutation of the colored blocks), we ensured that such trials were shown at least two experimental trials apart. 
To foster engagement and immersion in the task, the instructions and questions presented to participants incorporated the words ``I'' or ``my''. In the Action trials, participants were asked, ``What was my first action?'' whereas in the Goal trials, the participants were asked, ``What was the goal?'' To report their preferences, participants were provided with a slider bar featuring respective options on either side.

The attention check trials assessed the consistency in participants' preferences. A participant was considered consistent if all their preferences for a given trial fell on the same side of the slider bar. In accordance with this definition, participants who failed this consistency check for both attention check trials (which involved three variants created via color permutation of the blocks) were excluded from the analysis. Data from $83$ participants were collected via Prolific \cite{palan2018prolific}, of whom $23$ were excluded (based on the aforementioned exclusion criteria), yielding the pre-registered total of $60$ participants.

For the response time analysis, in addition to the aforementioned exclusion criteria, trials with response times $\pm 2$ standard deviations from the mean response time across all trials were excluded. This exclusion criterion was applied only to minimize the influence of outliers and ensure that anomalous response times did not bias our results.

\subsubsection*{Experiment 2: Explaining humans’ embodiment strategies (one- vs. two-hands)}
This experiment was also pre-registered \cite{shah2023}, including the stimuli, procedure, exclusion criteria, number of participants, and analysis. Participants rated the number of hands needed to rearrange an initial block configuration to a final block configuration. 

\textbf{Stimuli:} 
A total of $22$ trials were designed, each featuring a pair of images depicting initial and final block configurations (see \prettyref{suppfig:supp_stim_hands} for the full set of stimuli). Each stimulus comprised three to four blocks, all identical in shape—height twice the width—but differing in color, with blocks in yellow, blue, purple, and occasionally red. The stimuli were presented as images, with the ``Initial Configuration'' and ``Final Configuration'' blocks clearly labeled. 

\textbf{Procedure and Participants:} 
The $22$ trials were presented to the participants in random order. Just like in \hyperref[exp:1]{Experiment~1}, we employed a prose to give the impression that the experimenter had manipulated the block configurations presented to the participants. The instructions and questions addressed the experimenter using ``I'' or ``my''. In each experimental trial, the participants were asked, ``How many hands did I use?''. To report their preferences, participants were provided with a slider bar, ranging from ``Definitely just one hand'' on the left (corresponding to $0$ on the scale) to ``Definitely two hands'' on the right (corresponding to $100$ on the scale).
 
Three of the $22$ trials involved the movement of only a single block between the initial and final configurations. We used these trials for participant exclusion. We expected participants to comprehend that moving one block could only be accomplished using a single hand. Consequently, participants whose responses substantially deviated from the left side of the scale, i.e., whose responses were higher than $30$, were excluded from the analysis.

Data from $73$ participants were obtained via Prolific \cite{palan2018prolific}, but $13$ participants were excluded (based on the aforementioned exclusion criterion), giving us a pre-registered total of $60$ participants.

For the response time analysis, in addition to the aforementioned exclusion criteria, trials with response times $\pm 2$ standard deviations from the mean response time across all trials were excluded. This exclusion criterion was applied only to minimize the influence of outliers and ensure that anomalous response times did not bias our results.

\subsubsection*{Experiment 3: Generalizing beyond manipulation to navigation}

All details concerning the navigation experiment, including the stimuli, procedure, exclusion criteria, number of participants, and behavioral analysis, were pre-registered \cite{shah2026}.

\textbf{Stimuli:} 
We designed $22$ trials, each modeled in Blender \cite{blender521}, featuring three or more platforms of equal height ($3.5$ units) and featuring rounded corners. All platforms were positioned on a rigid, immovable ground surface with a dark wood texture, measuring $30 \times 8 \times .25$ units. The leftmost platform was always designated as the ``start'' platform, and the rightmost was always the ``final'' platform. Both platforms had widths twice their lengths, and were rendered to appear as though made of stone. A given stimulus could contain multiple intermediate platforms with the same shape and appearance. Across the stimuli, though, the intermediate platforms varied in shape, with dimensions ranging from $1.2$ to $8.0$ units, and/or in material composition, appearing as made of either wood or stone. All intermediate platforms used to build the stimuli set are shown in  \prettyref{suppfig:supp_nav_platforms}. For a subset of trials, we generated a replica in which the shapes and materials of the intermediate platforms remained unchanged, but their relative positions between the start and final platform were cyclically rotated. This manipulation allowed us to verify that judgments were driven by platform shape and material rather than by their spatial arrangement. Participants were shown only one exemplar from each such pair. The minimum gap between any two platforms was set to $0.25$ units. The top surfaces of the platforms were assigned a friction coefficient of $0.7$, whereas their bottom, in contact with the ground surface, was frictionless. An agent, approximately $2.3$ units tall, standing upright with its arms extended laterally, was shown standing on the start platform to provide participants with a visual reference for the relative scales of the agent and platforms.

The stimuli were first presented as videos in which the camera followed an elliptical trajectory around the scene. Beginning behind the agent, the camera first moved to reveal the left half of the scene, returned behind the agent, and then continued toward the right end before returning to a midpoint on the right side of the scene, where it came to rest with the start platform on the left and the final platform on the right. The participants were allowed to replay each video as many times as desired by pressing a designated key, but could not submit their judgment until the video had finished playing. At the end of each video, this final camera view was held on screen until participants reported their difficulty judgment.

Seven stimulus pairs were designed such that the position and shape of the platforms were identical within each pair, and only the material composition of one or more intermediate platforms was varied. Thus, each pair contrasted two trials: one in which the relevant intermediate platforms appeared to be made of stone and a corresponding trial in which the same platforms appeared to be made of wood.

\textbf{Procedure and Participants:} 
The $22$ trials were presented to the participants in random order. On each trial, participants were asked, ``How difficult would it be for the agent to reach the rightmost platform?''. Responses were collected using a slider bar ranging from 1 (leftmost) to 100 (rightmost), with increments of 1.

Data from $62$ participants were obtained via Prolific \cite{palan2018prolific}. Two participants were excluded for not completing the experiment within the allotted time, resulting in a final pre-registered total of $60$ participants.

\bibliography{sn-bibliography}


\section*{Acknowledgements}

We extend our heartfelt gratitude to everyone who provided feedback for this project, including members of the Cognitive \& Neural Computation Lab at Yale University, and to Marc Toussiant for his guidance on the KOMO framework.

\section*{Declarations}

\bmhead{Funding}
Not applicable

\bmhead{Conflict of interest/Competing interests}
The authors declare that they have no competing interests.

\bmhead{Ethics approval and consent to participate}
All the newly reported, preregistered behavioral experiments in our paper were approved by the Yale University Institutional Review Board, and each participant’s informed consent was obtained prior to any experimental sessions. 

\bmhead{Consent for publication}
Not applicable

\bmhead{Data, Materials and Code availability}
The source code for the computational models and behavioral experiments reported in this paper, the anonymized behavioral data, the experimental stimuli, and all the analysis scripts are available on GitHub at \url{https://github.com/CNCLgithub/genea}.

\bmhead{Author contribution}
A.D.S., B.S., T.G., and I.Y. designed manipulation-related behavioral experiments; A.D.S. and I.Y. designed navigation-related behavioral experiments; A.D.S. and I.Y. mathematically formulated models; A.D.S., B.S., J.B.T., T.G., and I.Y. designed computational implementation; A.D.S. performed data analysis. A.D.S. and I.Y. wrote the initial manuscript; A.D.S., J.B.T., T.G., H.G., and I.Y. edited the manuscript.


\clearpage
\backmatter

\setcounter{figure}{0}
\setcounter{page}{1}
\renewcommand{\figurename}{Supplementary Fig. }
\makeatletter
\renewcommand{\fnum@figure}{\textbf{\figurename\thefigure}}
\makeatother

\section*{Supplementary Information} \label{sec: supp_materials}


\begin{figure*}[tbhp]
\centering
\includegraphics[width=\linewidth]{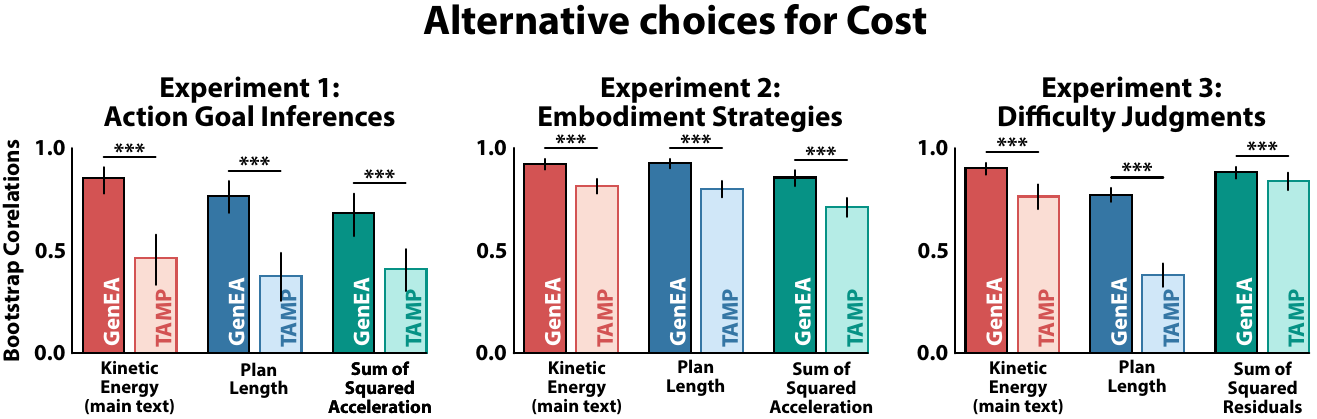}
\let\nobreakspace\relax
\caption{GenEA is robust to the choice of the cost function that parametrizes the distribution over plans, Pr$(\pi\,|\,R,\,\xi)$. This figure shows bootstrapped correlations between GenEA and human performance across all experiments we reported in the main text, and compares to standard TAMP. Irrespective of the function we used to compute the cost, the GenEA model outperformed TAMP-based agent representation. Error bars in this figure indicate $95\%$ bootstrapped confidence intervals; "***" for $p < .001$.}
\label{suppfig:supp_cost_funcs}
\end{figure*}


\clearpage


\begin{figure*}
\centering
\includegraphics[width=0.8\linewidth]{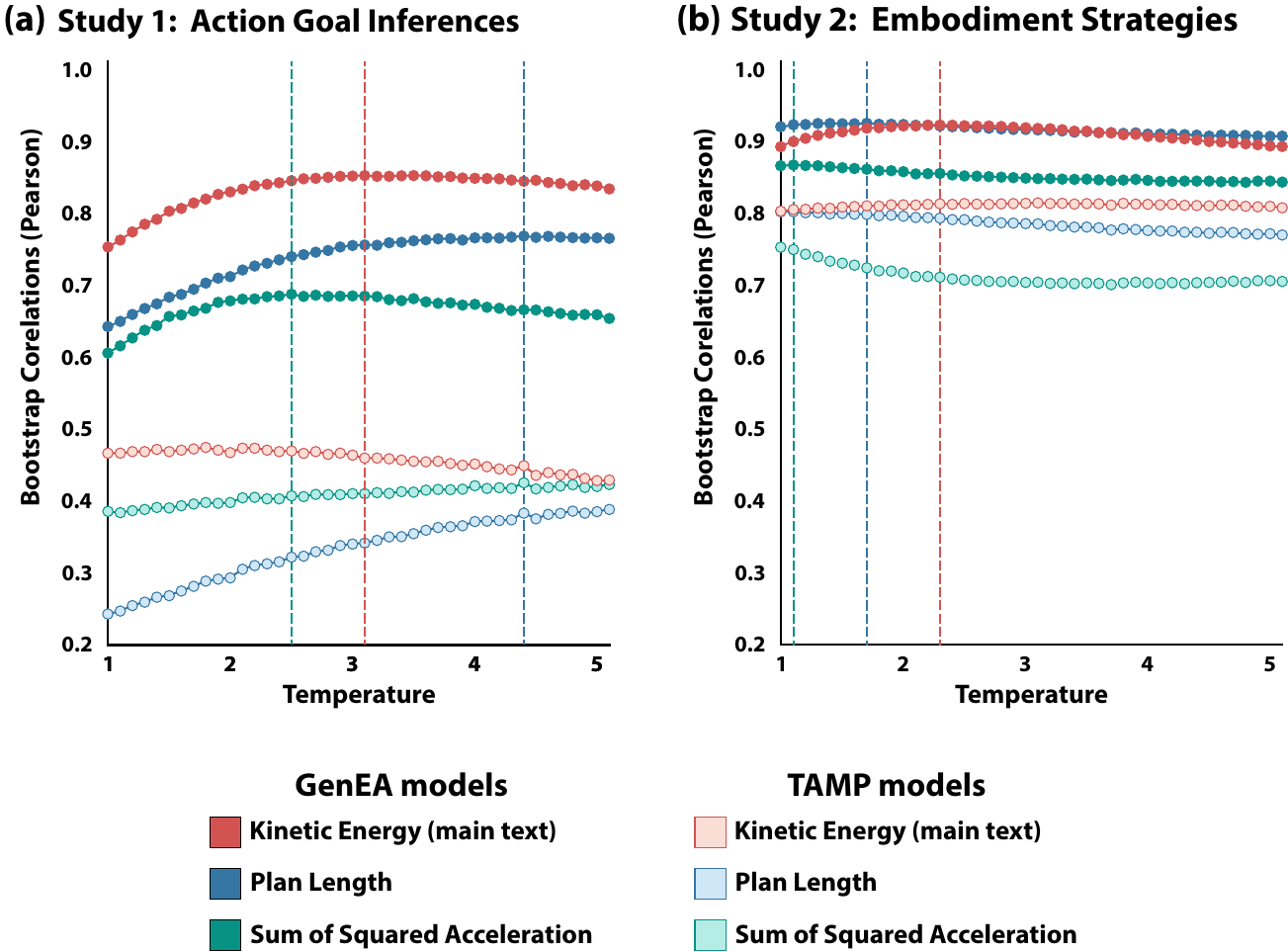}
\let\nobreakspace\relax
\caption{Our results are not sensitive to the choice of the temperature parameter $\kappa$. This plot shows that for each cost function we considered, and across all temperature values (x-axis), GenEA consistently outperforms the corresponding TAMP-based model. The dotted lines in the plot indicate the temperature values we used for each cost function to generate the results shown in \prettyref{suppfig:supp_cost_funcs}: in \hyperref[exp:1]{Experiment 1}, $\kappa = 3.1$ for Kinetic energy (also the value we used in main text), $\kappa = 4.4$ for plan length, and $\kappa = 2.5$ for sum of squared errors; and in \hyperref[exp:2]{Experiment 2}, , $\kappa = 3.5$ for Kinetic energy (also the value we used in main text), $\kappa = 2.4$ for plan length, and $\kappa = 1.1$ for sum of squared errors}
\label{suppfig:supp_temp_prob_pearson}
\end{figure*}


\clearpage


\begin{figure*}
\centering
\includegraphics[width=.5\linewidth]{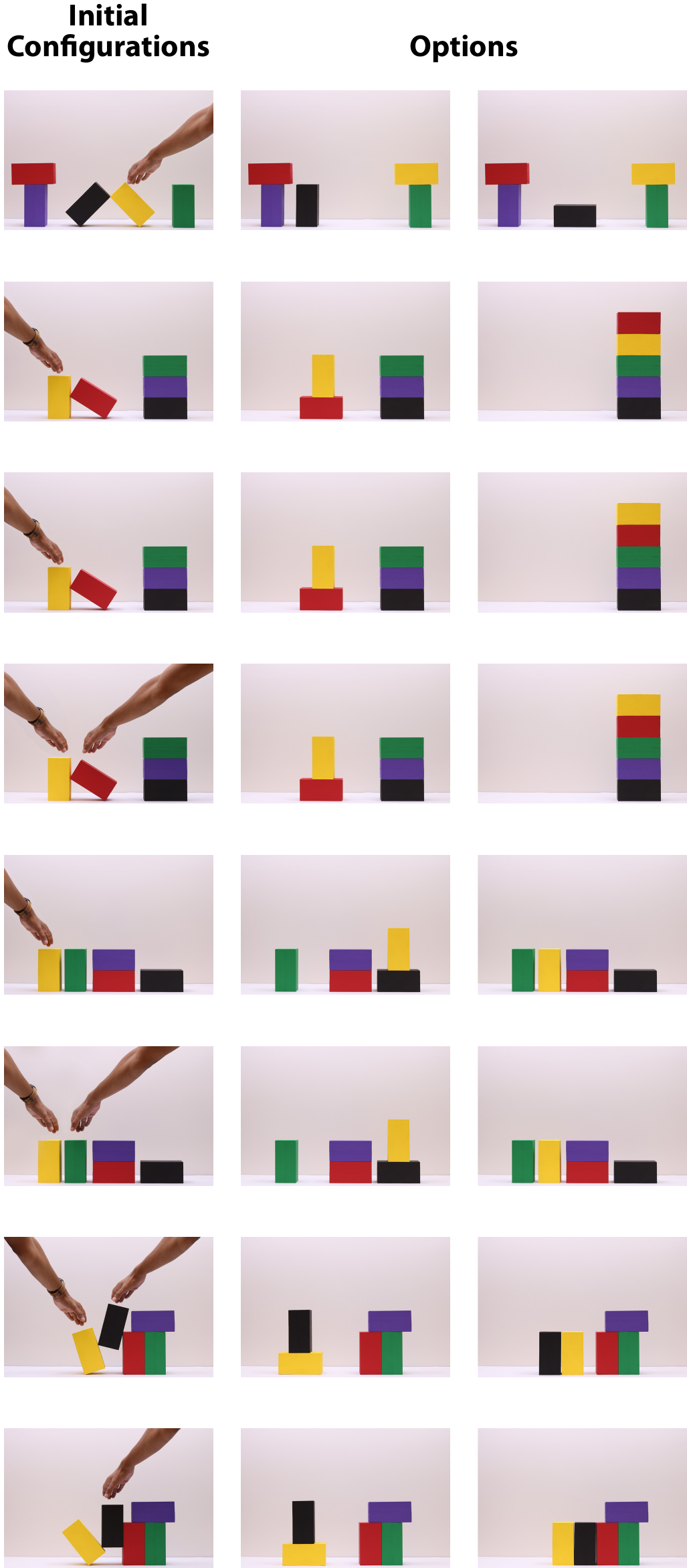}
\let\nobreakspace\relax
\caption{All goal inference stimuli used in \hyperref[exp:1]{Experiment 1}.}
\label{suppfig:supp_stim_social_goal}
\end{figure*}


\clearpage


\begin{figure*}
\centering
\includegraphics[width=.6\linewidth]{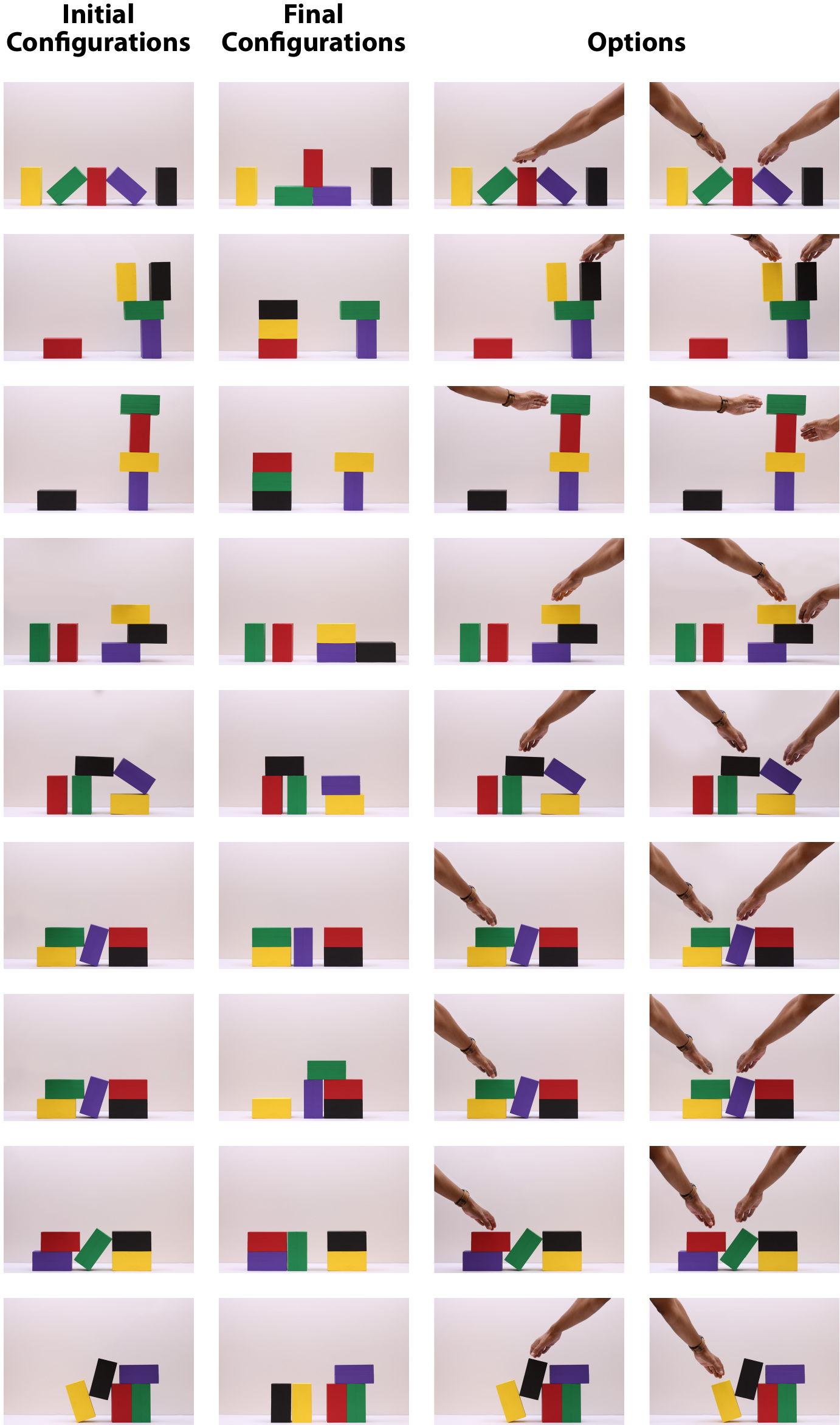}
\let\nobreakspace\relax
\caption{All action inference stimuli used in \hyperref[exp:1]{Experiment 1}.}
\label{suppfig:supp_stim_social_action}
\end{figure*}


\clearpage


\begin{figure*}
\centering
\includegraphics[width=.6\linewidth]{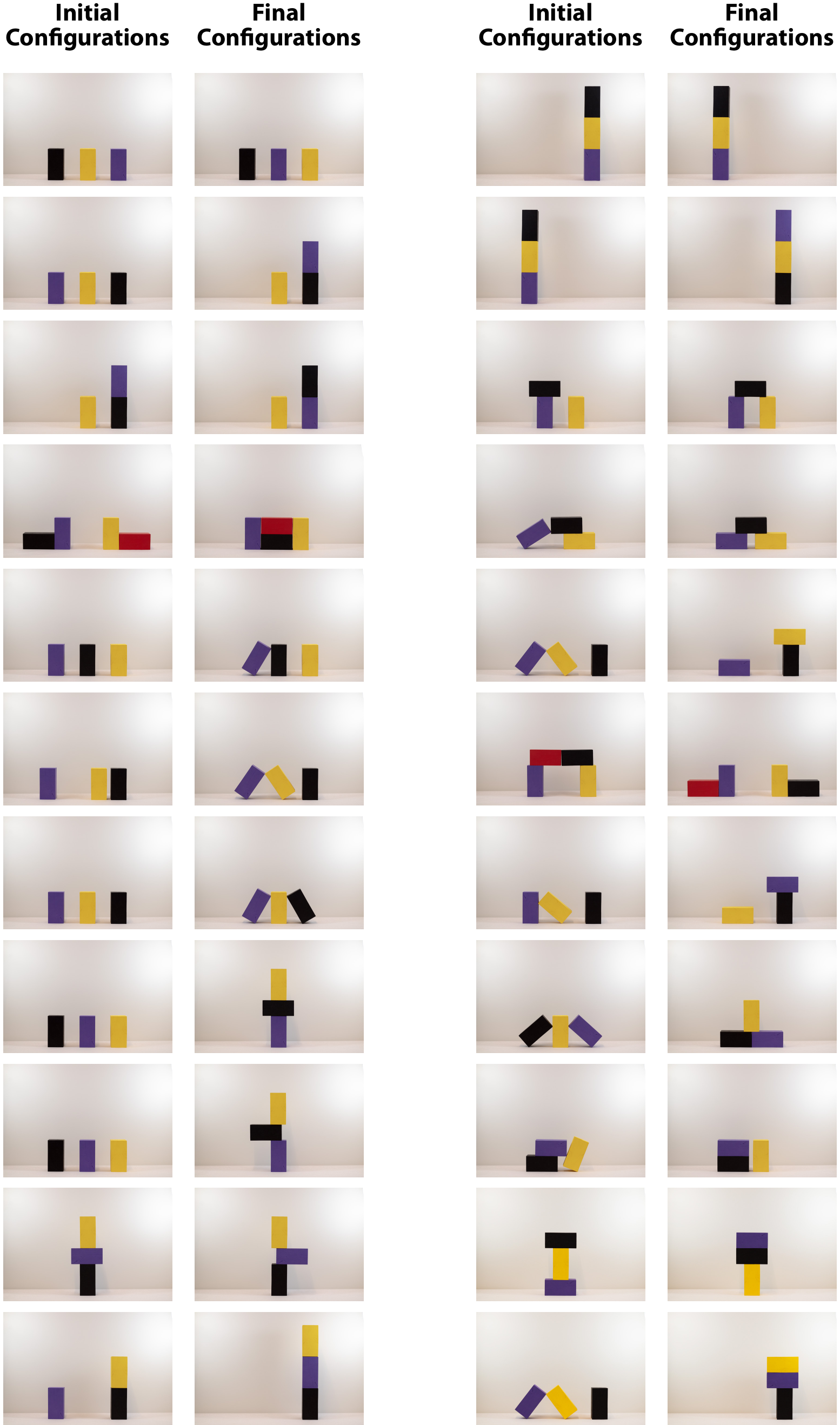}
\let\nobreakspace\relax
\caption{All stimuli used in \hyperref[exp:2]{Experiment 2}.}
\label{suppfig:supp_stim_hands}
\end{figure*}


\clearpage


\begin{figure*}
\centering
\includegraphics[width=\linewidth]{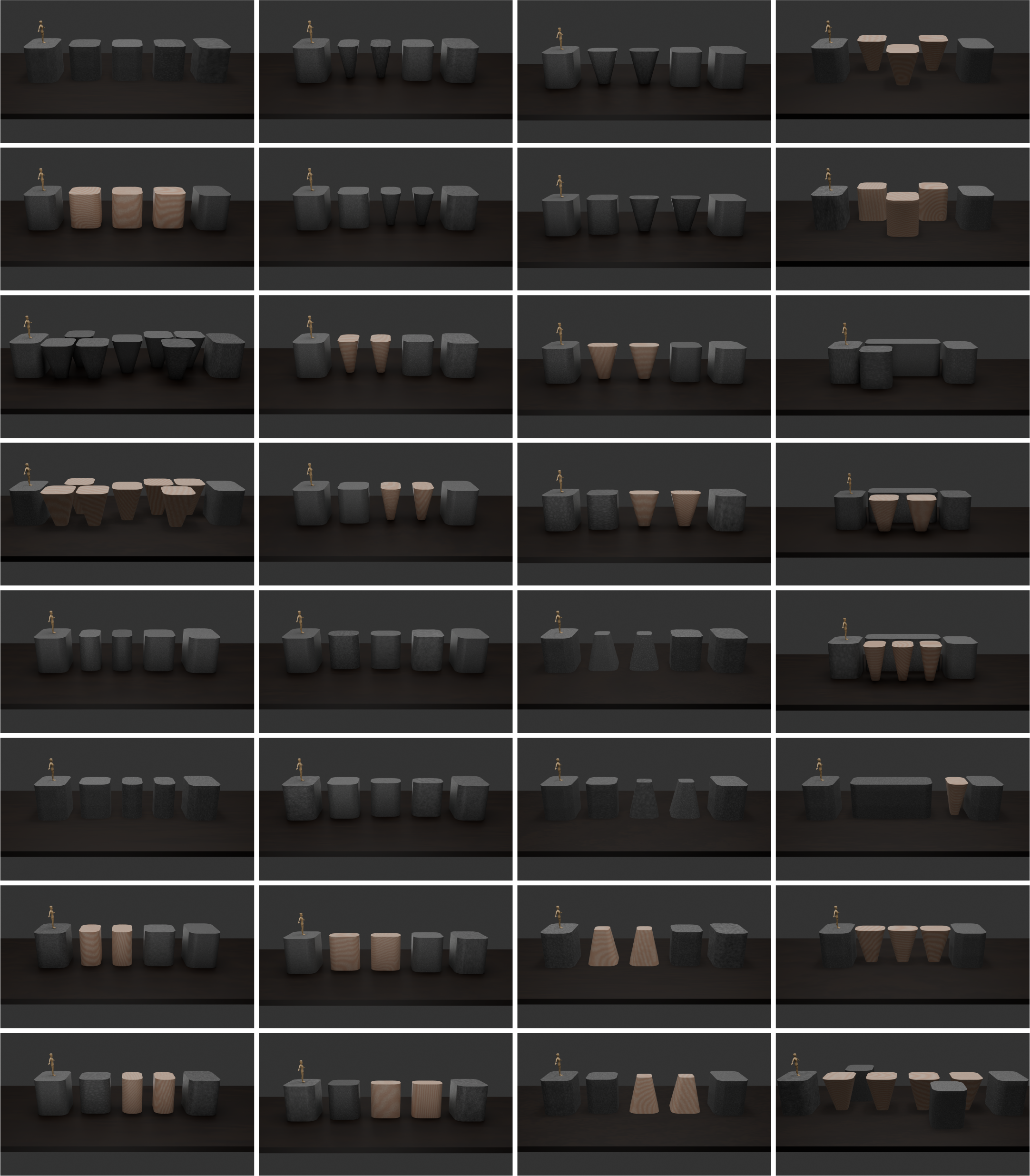}
\let\nobreakspace\relax
\caption{The stimuli set from \hyperref[exp:3]{Experiment 3}.}
\label{suppfig:supp_stim_diff}
\end{figure*}


\clearpage


\begin{figure*}
\centering
\includegraphics[width=.6\linewidth]{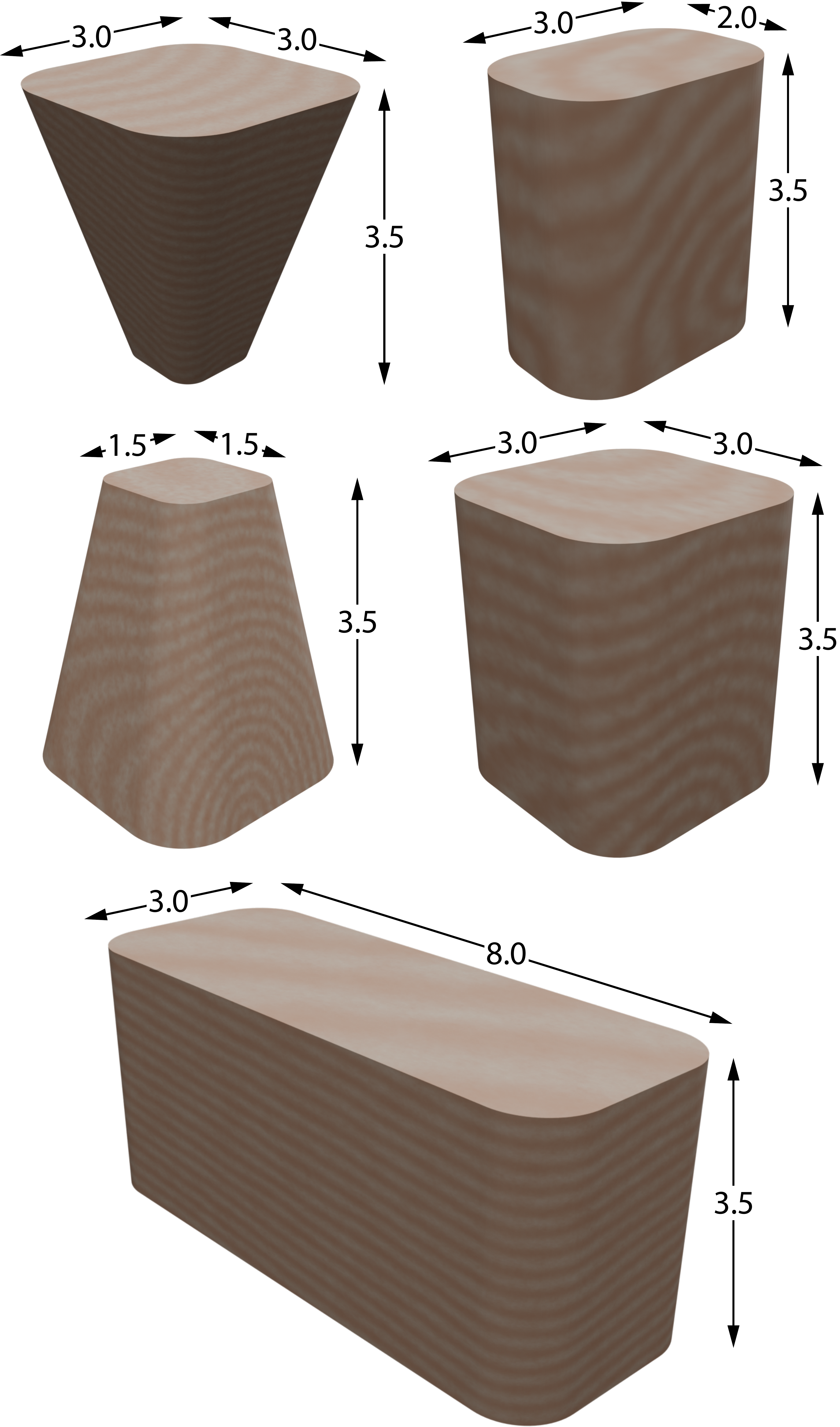}
\let\nobreakspace\relax
\caption{Platforms used in the \hyperref[exp:3]{Experiment 3} stimuli set, along with their dimensions. The platforms are depicted with a wooden material composition, but in the actual stimuli set they could also appear as though made of stone.}
\label{suppfig:supp_nav_platforms}
\end{figure*}


\clearpage


\begin{figure*}
\centering
\includegraphics[width=.95\linewidth]{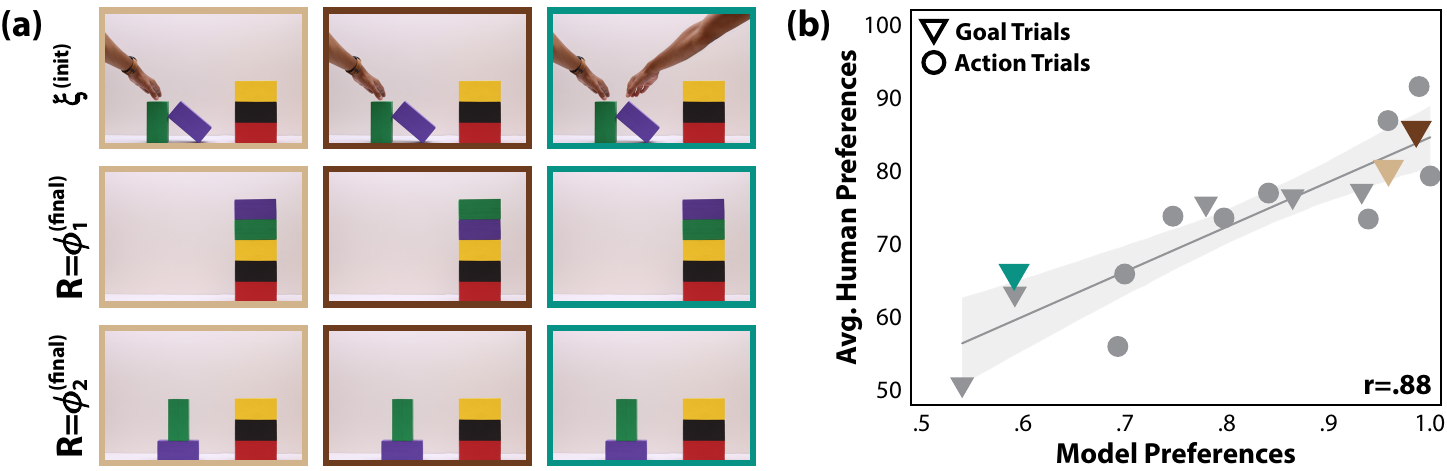}
\let\nobreakspace\relax
\caption{
\textbf{(a)} Each column shows example goal inference trials from \hyperref[exp:1]{Experiment 1}. In each column, the top row displays the initial block configuration $\phi^{(\text{init})}$, the second and third rows show the reward query $Q_R=\{\phi^{(\text{final})}_1,\,\phi^{(\text{final})}_2\}$. Note that $Q_{\pi}=\{\xi^{(\text{init})}\}$. 
\textbf{(b)} We use this caption to provide qualitative insight as to how GenEA aligns with human judgments. The scatter plot is reproduced from the main text (Fig. 4) for reader's reference, with the colored dots corresponding to the trials shown on panel (a). As a reminder, for both the participants and the model, small values indicate preference toward the more optimal goal, whereas larger values indicate ambivalence between the two options.  
Here, the only difference between the ``golden'' (left-most column) and ``brown'' (middle column) trials is in their respective final configurations $\phi^{(\text{final})}_1$ --- the positions of the top two blocks in the stack are swapped between the two reward configurations. As a result, it is only in the ``golden'' trial that the final location of the green block, picked up during the first action, is immediately available. The results suggest that the judgments of both the human participants as well as GenEA are affected by this disparity. While they both show a strong preference for $\phi^{(\text{final})}_2$, the presence of a relatively less plausible alternative in the ``golden'' trial (i.e., $\phi^{(\text{final})}_1$) introduces greater ambiguity for this trial as compared to the ``brown'' trial. The ``pink'' trial is identical to the ``golden'' trial in every way except the initial action taken by the agent. With a two-handed action available in the ``pink'' trial, both options become equally plausible, and we see that the preferences of the human participants as well as GenEA strongly reflect this ambivalence.}
\begingroup
\phantomsubcaption
\label{suppfig:supp_social_goalA}
\phantomsubcaption
\label{suppfig:supp_social_goalB}
\phantomsubcaption
\label{suppfig:supp_social_goalC}
\endgroup
\label{suppfig:supp_social_goal}
\end{figure*}


\clearpage


\begin{figure*}
\centering
\includegraphics[width=\linewidth]{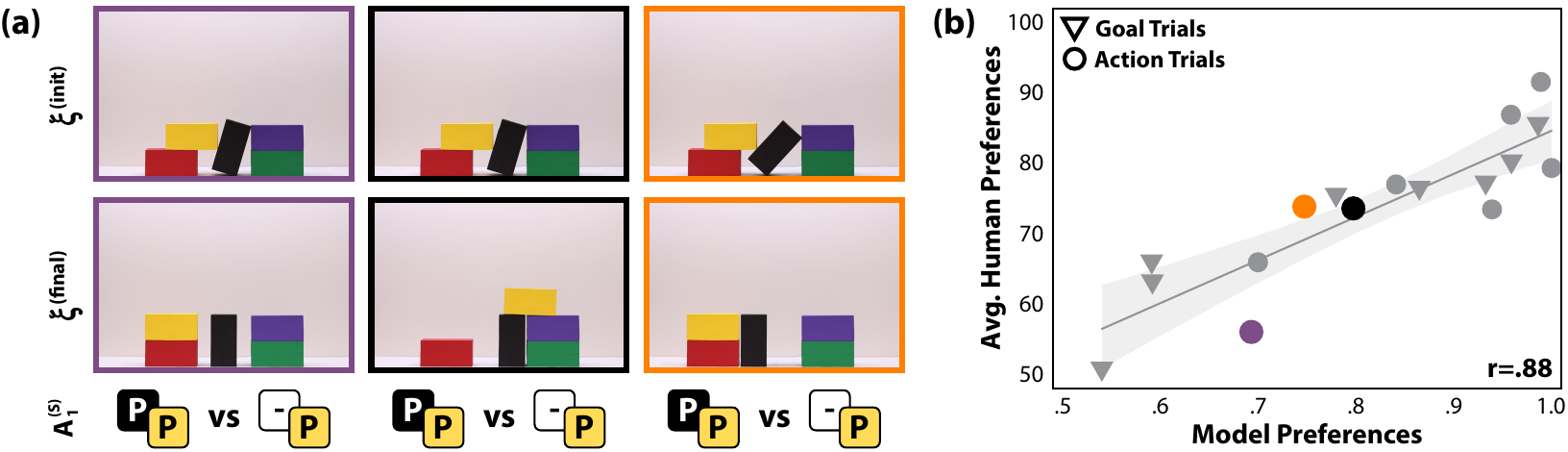}
\let\nobreakspace\relax
\caption{
\textbf{(a)} Each column shows an example action inference trials from \hyperref[exp:1]{Experiment 1}. In each column, the top row displays an initial configuration $\phi^{(\text{init})}$, the middle row shows a final configuration $Q_R=\{\phi^{(\text{final})}\}$, and the third row shows the two possible first actions, from the two query constraints $Q_{\pi}=\{\xi^{(\text{constr})}_1,\,\xi^{(\text{constr})}_2\}$, the agent may take on the initial block configuration. Notice that for all three example trials shown in this figure, the action options are identical: \textit{(i)} a two-handed action involving picking the black block with the left hand and picking the yellow block with the right hand; \textit{(ii)} a one-handed action involving picking the yellow block with the right hand (i.e., no action is scheduled on the left hand).  
\textbf{(b)} The scatter plot, reproduced from the main text (Fig. 4) for reference, with the colored dots corresponding to the trials shown on panel (a). As a reminder, for both the participants and the model, larger values on either axis indicate preference toward the more optimal action, whereas smaller values indicate increasing ambivalence between the two action options. The ``purple'' and the ``black'' trials have the same initial conditions but differ markedly in their final configurations. In the ``purple'' trial, merely moving the yellow block allows the black block to stand upright (due to gravity), and achieve the final position in the process. Consequently, for the ``purple'' trial, both humans and GenEA showed some ambivalence with only a slight preference for the one-handed action which leverages gravity. In contrast, for the ``black'' trial, both humans and GenEA strongly preferred the more efficient two-handed action. The ``orange'' trial introduces a more nuanced variation to the ``purple'' trial with the only difference being the tilt angle of the black block in the initial configuration. Unlike the ``purple'' trial, where gravity can assist the black block in reaching its final position, the ``orange'' trial features a larger tilt angle, necessitating a two-handed action to accomplish the task efficiently. We see in the plot that, in this trial as well, both humans and GenEA exhibited a strong preference for the two-handed action.
}
\begingroup
\phantomsubcaption
\label{suppfig:supp_social_actionA}
\phantomsubcaption
\label{suppfig:supp_social_actionB}
\phantomsubcaption
\label{suppfig:supp_social_actionC}
\endgroup
\label{suppfig:supp_social_action}
\end{figure*}


\clearpage


\begin{figure*}
\centering
\includegraphics[width=\linewidth]{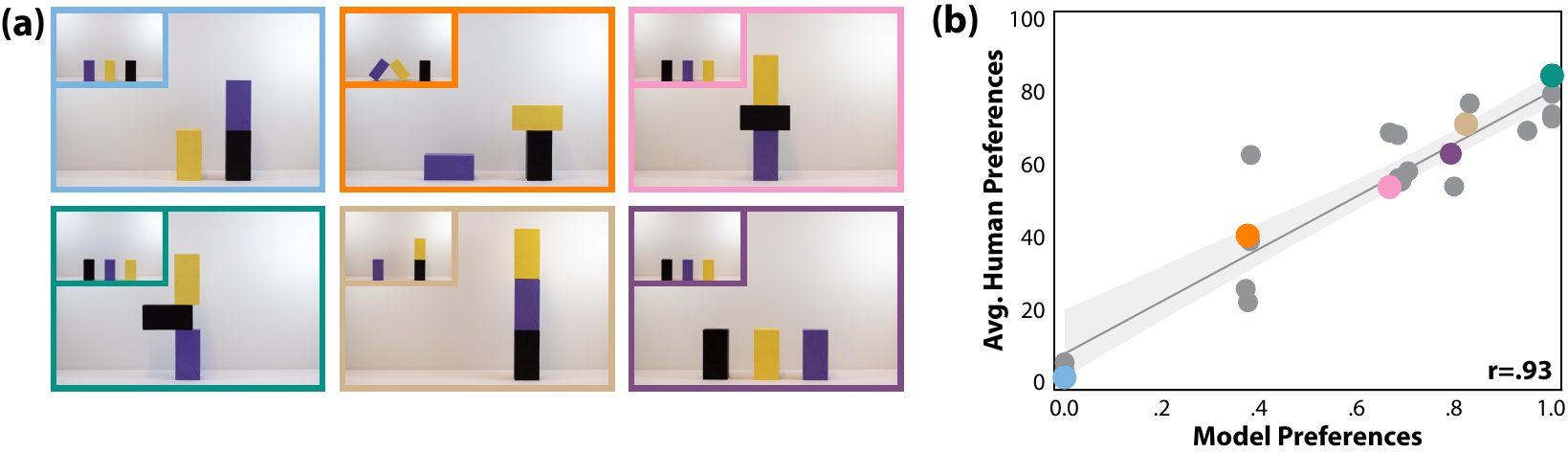}
\let\nobreakspace\relax
\caption{
\textbf{(a)} Each image shows an example trial in \hyperref[exp:2]{Experiment 2}, with an initial configuration (inset) $\phi^{(\text{init})}$ and final configuration $Q_R = \{\phi^{(\text{final})}\}$. The two query constraints are $Q_{\pi}=\{\xi^{(\text{constr})}_1,\,\xi^{(\text{constr})}_2\}$, plans using one-hand and plans using two-hands, respectively. 
\textbf{(b)} The scatter plot, reproduced from the main text (Fig. 5) for reference, where the colored dots correspond to the trials shown on panel (a). As a reminder, for both the participants and the model, smaller values imply preference toward one-hand, whereas larger values imply preference toward two-hand.
The cluster of points in the lower left corner (e.g., the ``blue'' trial) represent trials where only one block was moved in the initial-final configuration pair, and both humans and GenEA show preference for one-hand. 
The trials in the top-right corner also demonstrate a high degree of alignment between humans and GenEA. These trials involve configurations where the blocks' support structure necessitates using both hands during their construction. Take, for instance, the ``green'' trial where the black block cannot be held in its place without the yellow block's support. To prevent the black block from toppling, the yellow block must be ``Fixed'' simultaneously. 
In some ways, it is the trials situated toward the center of the plot that most clearly reveal GenEA’s strengths. Typically, a two-hand plans are more efficient than their one-handed counterparts: they involve fewer moves and incur lower execution costs. Such trials are displayed in the right half (for the model) and upper half (for humans) of the scatter plot. In the ``golden'' trial, for instance, which involves swapping the positions of two vertically stacked blocks, a two-handed solution is clearly more efficient and therefore also more preferable. This advantage is less pronounced in the ``purple'' trial, which, like the ``golden'' trial, also involves swapping but presents fewer positional constraints since the blocks to be swapped are positioned on the table. In the ``pink'' trial, whose final configuration can be conveniently achieved using either one-handed or two-handed solutions, both humans as well as GenEA show the most ambivalence, with only a slight preference for the more efficient two-handed solution. Our stimuli set also includes trials where leveraging external forces, such as gravity, can make one-handed plans more efficient. For instance, in the ``orange'' trial, moving the yellow block causes the black block to fall into its final position. Such trials lie in the lower-left quadrant of the plot.}
\begingroup
\phantomsubcaption
\label{suppfig:supp_handsA}
\phantomsubcaption
\label{suppfig:supp_handsB}
\endgroup
\label{suppfig:supp_hands}
\end{figure*}


\clearpage


\begin{figure*}
\centering
\includegraphics[width=\linewidth]{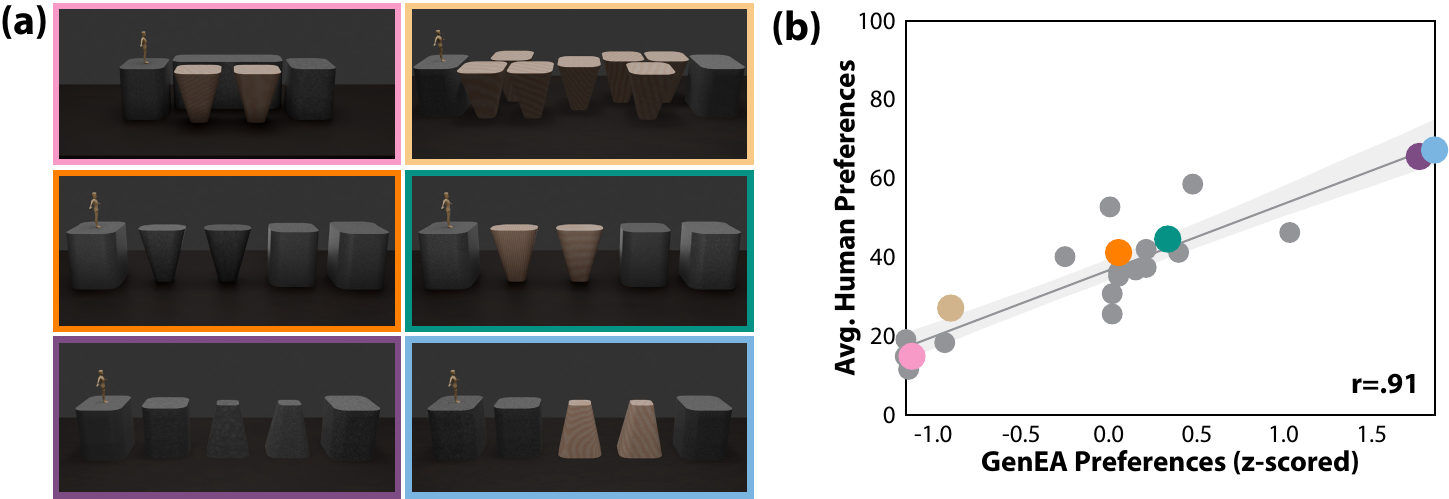}
\let\nobreakspace\relax
\caption{
\textbf{(a)} Each image shows the initial agent and object configuration for a representative trial from \hyperref[exp:3]{Experiment 3}. 
\textbf{(b)} We use this caption to provide qualitative insight into how GenEA's judgments aligns with those of human participants in \hyperref[exp:3]{Experiment 3}. The scatter plot, reproduced from the main text (\prettyref{fig:diff_nav}), shows colored dots corresponding to the trials depicted in panel (a). As a reminder, smaller values indicate easier trials for the participants (and lower cost under GenEA), while higher values indicate harder trials.
Both humans and GenEA consistently judge configurations that can be traversed by walking as easier than those requiring jumps. Similarly, they are both sensitive to the material composition of the platforms, judging configurations with wooden platforms as more difficult than otherwise identical configurations with stone platforms, as is evident in the orange versus green and purple versus blue trial pairs. The gold trial illustrates that platform count and material alone do not determine task difficulty. Although this trial contains the most intermediate platforms, all wooden and tapered at the base in a manner that renders them precarious, their close positioning affords a walkable path, and the trial is accordingly judged among the least difficult to traverse. Both humans and GenEA further show that, in configurations admitting multiple candidate paths, difficulty judgments implicitly depend on first inferring which path the agent would take. In the pink trial, for example, a single long, stable platform affords a walkable route, and both humans and GenEA ground their difficulty judgments in this route rather than in the two comparatively unstable platforms in the foreground. Accordingly, both humans and GenEA judge the pink trial among the easiest. The blue and purple trials offer only a small landing area and as a result are regarded as the most difficult configurations by both humans and GenEA. Notably, this agreement between humans and GenEA is not confined to the extremes of the difficulty scale; but extends to configurations of intermediate difficulty as well.
}
\label{suppfig:supp_diff}
\end{figure*}


\clearpage


\begin{figure*}
\centering
\includegraphics[width=\linewidth]{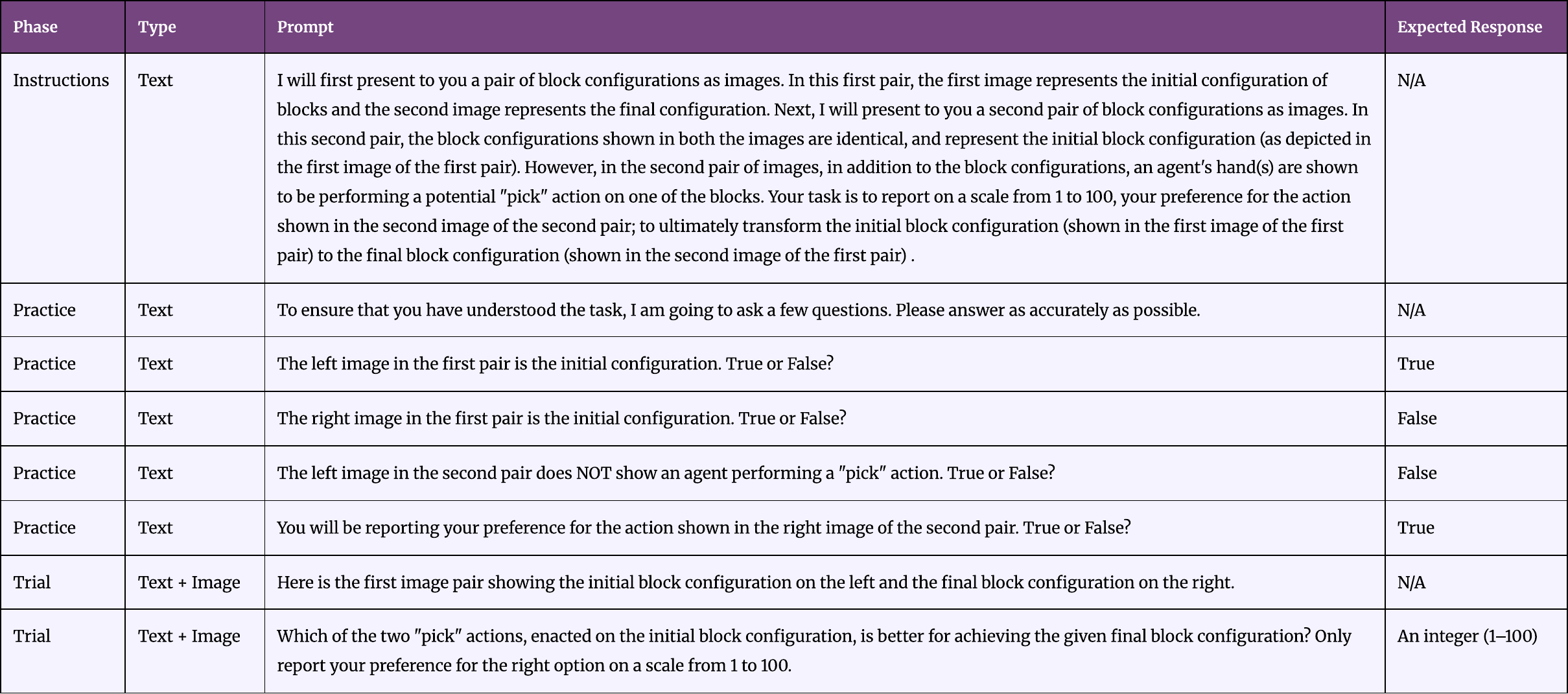}
\let\nobreakspace\relax
\caption{VLM prompts used for the action trials of \hyperref[exp:1]{Experiment 1}.}
\label{suppfig:supp_vlm_exp_action}
\end{figure*}


\clearpage


\begin{figure*}
\centering
\includegraphics[width=\linewidth]{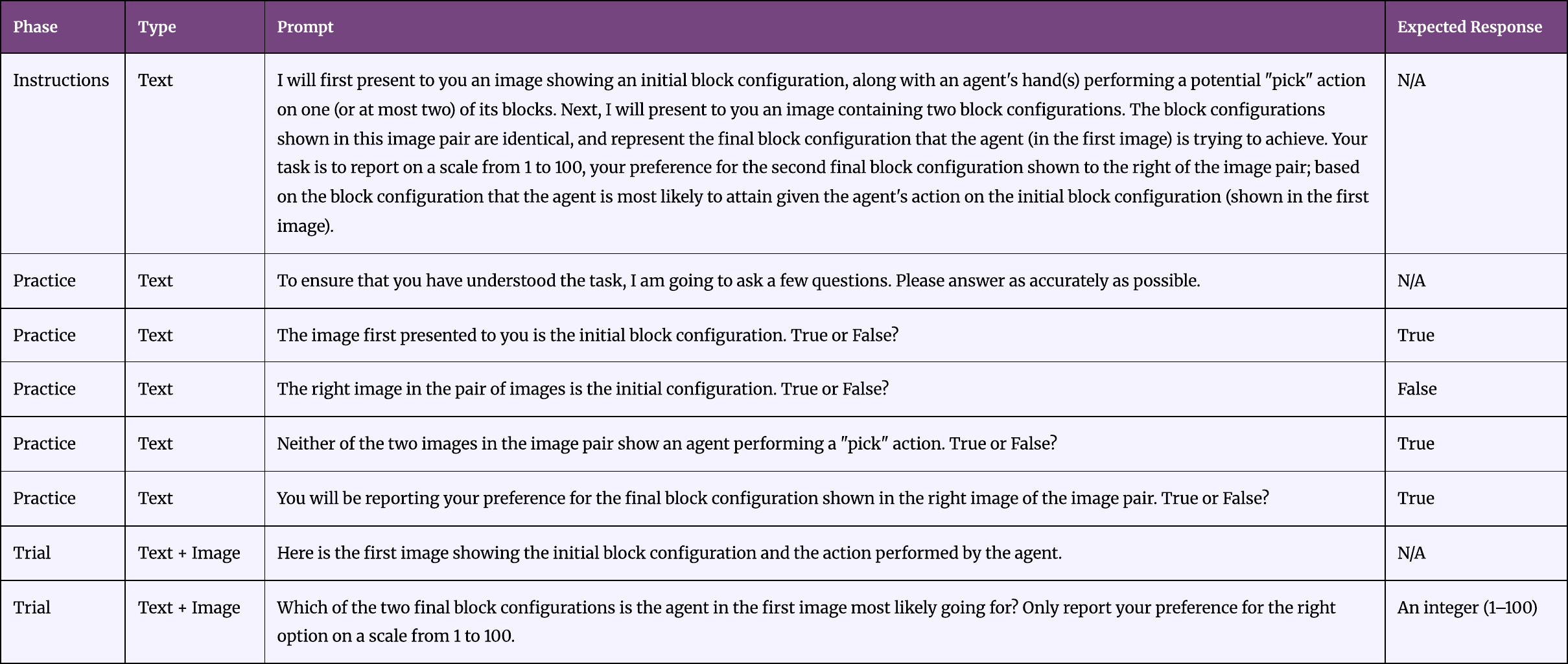}
\let\nobreakspace\relax
\caption{VLM prompts used for the goal trials of \hyperref[exp:1]{Experiment 1}.}
\label{suppfig:supp_vlm_exp_goal}
\end{figure*}


\clearpage


\begin{figure*}
\centering
\includegraphics[width=\linewidth]{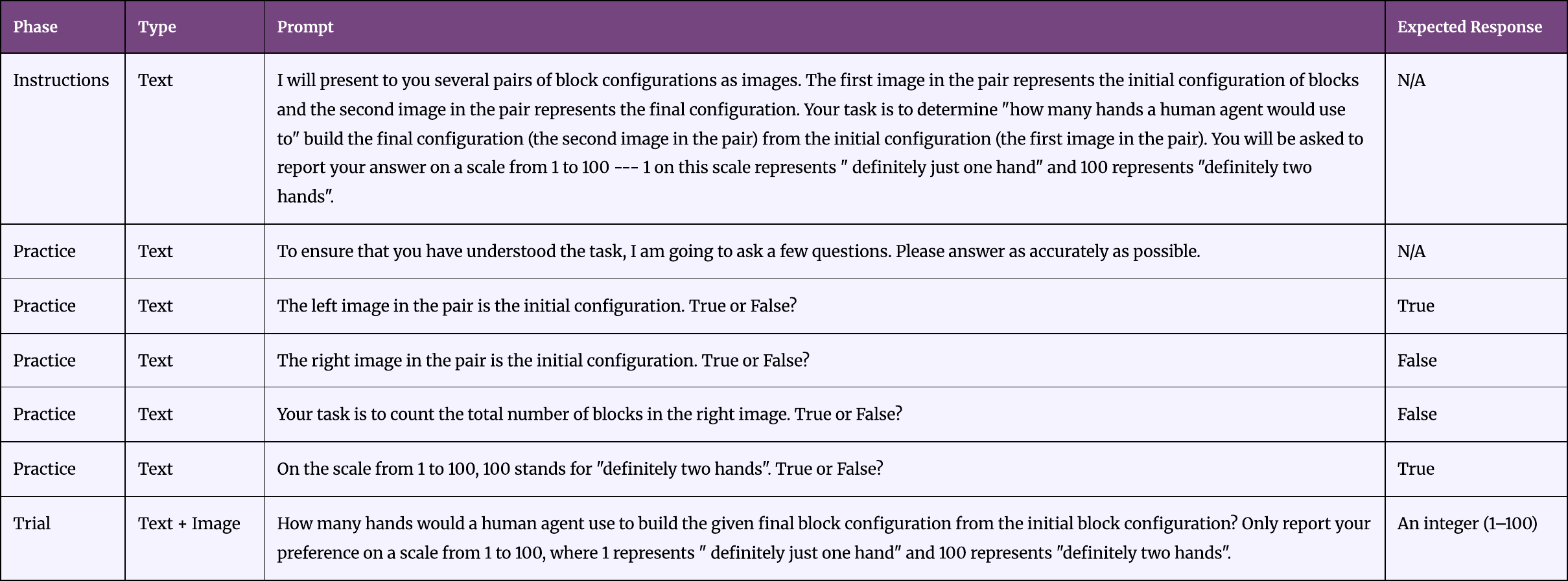}
\let\nobreakspace\relax
\caption{VLM prompts used for \hyperref[exp:2]{Experiment 2}.}
\label{suppfig:supp_vlm_exp_hands}
\end{figure*}


\clearpage


\begin{figure*}
\centering
\includegraphics[width=\linewidth]{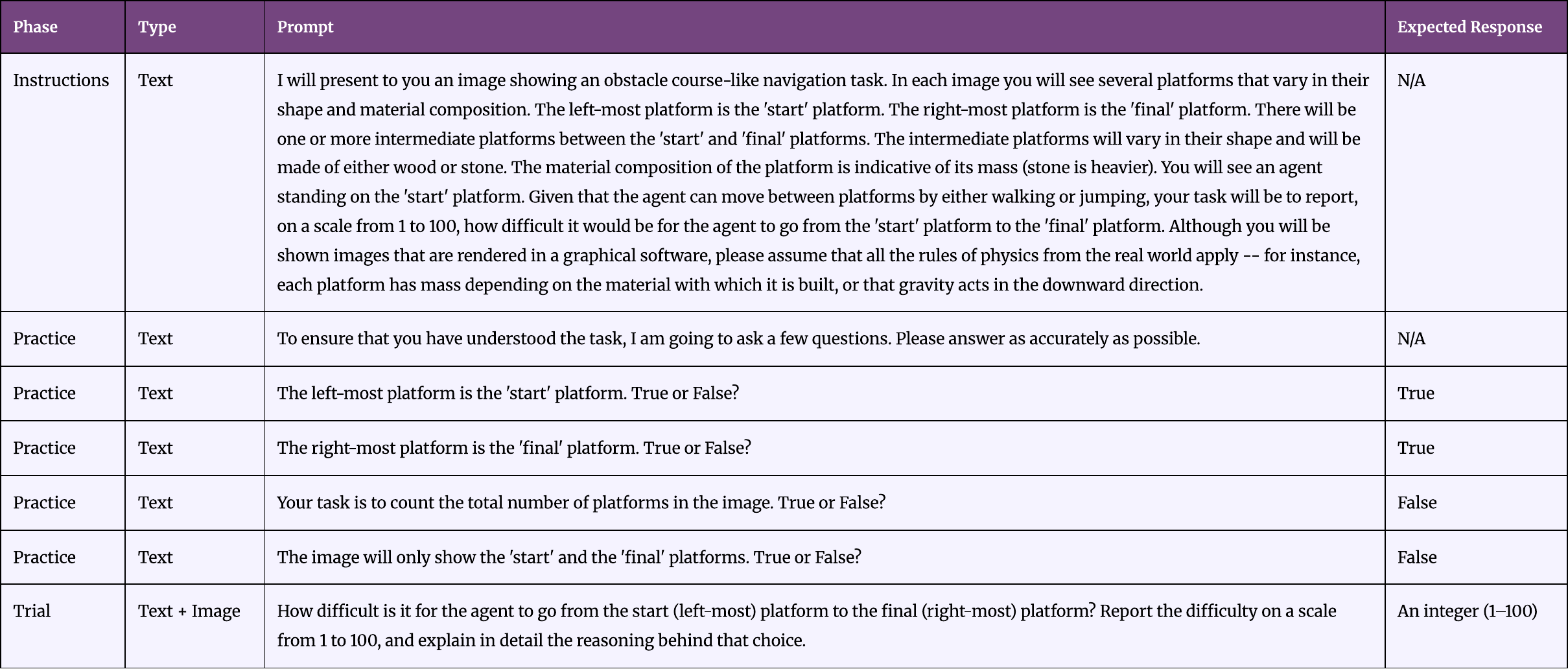}
\let\nobreakspace\relax
\caption{VLM prompts used for trials in \hyperref[exp:3]{Experiment 3}.}
\label{suppfig:supp_vlm_exp_nav_diff}
\end{figure*}


\end{document}